\ifdefined\XeTeXversion\else\pdfoutput=1\fi
\PassOptionsToPackage{most,breakable}{tcolorbox}
\PassOptionsToPackage{dvipsnames,table}{xcolor}
\PassOptionsToPackage{numbers,compress}{natbib}
\documentclass[11pt]{article}

\usepackage[preprint]{ropedia}

\usepackage[T1]{fontenc}
\usepackage[utf8]{inputenc}
\usepackage{amsmath,amssymb}
\usepackage{amsfonts}
\usepackage{bm}
\usepackage{mathpazo}
\usepackage[xcharter,bigdelims,vvarbb]{newtxmath}
\usepackage[scaled]{helvet}
\usepackage[scaled=1.1]{zlmtt}
\usepackage{pifont}
\usepackage{placeins}
\usepackage{booktabs}
\usepackage{multirow}
\usepackage{tabularx}
\usepackage{subcaption}
\usepackage{makecell}
\usepackage{enumitem}
\usepackage{titletoc}
\usepackage{needspace}
\usepackage{nicefrac}
\usepackage{microtype}
\usepackage{graphicx}
\usepackage{xspace}
\setlist[itemize]{leftmargin=2em}

\usepackage{tikz}
\usetikzlibrary{fadings}
\usepackage{array}
\usepackage{tcolorbox}
\usepackage{threeparttable}

\newcommand{\projecttitle}{DONIC}
\newcommand{\projectsubtitle}{Generalist Dexterous Bimanual Tracking Control from Human Demonstrations}

\definecolor{subcol}{HTML}{F7FFF2}
\colorlet{rankfirst}{ropediagreen}
\colorlet{ranksecond}{ropedialightgreen}
\colorlet{rankthird}{ropediaabstractgreen}
\colorlet{bestgreen}{rankfirst}
\colorlet{secondgreen}{ranksecond}
\colorlet{thirdgreen}{rankthird}
\colorlet{bestred}{rankfirst}
\colorlet{secondorange}{ranksecond}
\colorlet{thirdyellow}{rankthird}
\definecolor{trainmark}{RGB}{0,180,0}
\definecolor{oomred}{RGB}{252,228,228}
\definecolor{catgray}{HTML}{E7F8DC}
\definecolor{altcolgreen}{HTML}{F0FFE8}
\colorlet{altcolblue}{altcolgreen}
\definecolor{pretrainmark}{RGB}{197,90,17}
\definecolor{gaingreen}{RGB}{30,140,60}
\definecolor{lossred}{RGB}{200,30,30}
\definecolor{navyblue}{HTML}{4F8F35}

\colorlet{findingback}{ropedialightgreen}
\colorlet{findingbacksoft}{ropediaabstractgreen}
\colorlet{findingframe}{ropediadeepgreen}
\colorlet{findingaccent}{ropediaaccent}

\usepackage{amsmath,amsfonts,bm}

\def\eqref#1{equation~\ref{#1}}

\def\1{\bm{1}}

\def\va{{\bm{a}}}

\def\vo{{\bm{o}}}
\def\vp{{\bm{p}}}
\def\vq{{\bm{q}}}

\def\vs{{\bm{s}}}

\def\vv{{\bm{v}}}

\def\mR{{\bm{R}}}

\DeclareMathAlphabet{\mathsfit}{\encodingdefault}{\sfdefault}{m}{sl}
\SetMathAlphabet{\mathsfit}{bold}{\encodingdefault}{\sfdefault}{bx}{n}

\newcommand{\E}{\mathbb{E}}

\newcommand{\R}{\mathbb{R}}

\newcommand{\method}{\textsc{DONIC}}
\newcommand{\RopediaAuthor}[1]{\mbox{\textbf{#1}\textsuperscript{\ropediamark}}}
\newcommand{\RopediaAuthorCorr}[1]{\mbox{\textbf{#1}\textsuperscript{\ropediamark,$\dagger$}}}

\firstpagelogos{logo/ropedia_Green.png}{}{}
\abstractlogo{logo/Horizontal_Black.png}

\title{\projecttitle}
\subtitle{\projectsubtitle}

\author{
  \RopediaAuthor{Mingze Sun}\hspace{1.1em}
  \RopediaAuthor{Ziang Cao}\hspace{1.1em}
  \RopediaAuthor{Yuan Fei}\hspace{1.1em}
  \RopediaAuthor{Qiankuan Tao}\hspace{1.1em}
  \RopediaAuthor{Zihan Wang}\hspace{1.1em}
  \RopediaAuthor{Tianqi Liu}\hspace{1.1em}
  \RopediaAuthor{Junhao Chen}\hspace{1.1em}
  \RopediaAuthor{Haiwen Diao}\hspace{1.1em}
  \RopediaAuthor{Fangzhou Hong}\hspace{1.1em}
  \RopediaAuthor{Zhaoxi Chen}\hspace{1.1em}
  \RopediaAuthorCorr{Ziwei Liu}\\[1.0ex]
  {\normalfont\footnotesize \ropediamark\,Ropedia}
}

\begin{document}
\maketitle
{\renewcommand{\thefootnote}{}\footnotetext{$\dagger$ Corresponding author.}\addtocounter{footnote}{-1}}

\begin{paperresources}
\paperresourceicon{\resourceprojecticon}{Project Page}{https://ropedia.github.io/DONIC/}{ropedia.github.io/DONIC}
\end{paperresources}

\vspace{6pt}
{\centering
\includegraphics[width=\textwidth]{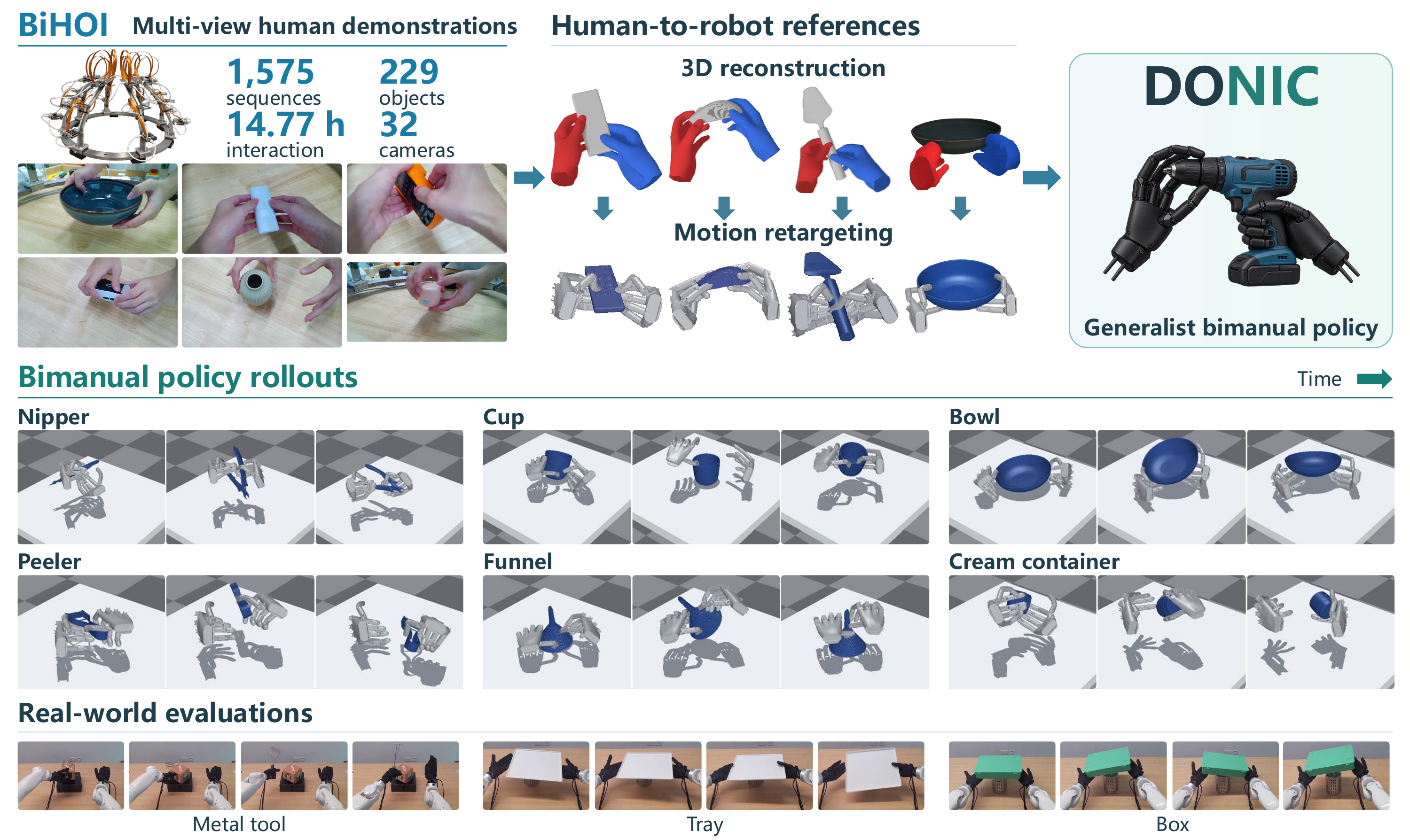}
\par\vspace{5pt}
\begin{minipage}{\textwidth}
\footnotesize
\refstepcounter{figure}\label{fig:teaser}
Figure~\thefigure{} | \textbf{Overview of \method.}
    BiHOI provides multi-view bimanual demonstrations that are
    reconstructed and retargeted to robot hands.
    \method\ learns a shared policy for diverse bimanual
    reference-tracking tasks.
\end{minipage}
\par}
\vspace{6pt}

\begin{abstract}
Despite recent progress in dexterous reference tracking, existing models focus primarily on single-hand manipulation and a limited range of interactions, leaving the bimanual dexterous behaviors common in real life unaddressed. The scarcity of contact-rich bimanual manipulation data and the increased complexity of coordinating contacts make developing a generalist bimanual tracker particularly challenging. Therefore, we present \textbf{1) \method}, the first generalist bimanual tracker jointly trained across operation categories and human-motion sources, together with \textbf{2) BiHOI}, a multi-view dataset of $1{,}575$ sequences spanning $229$ objects and emphasizing sustained bimanual interaction, where reconstructed interactions are retargeted into robot references. \method\ combines proximal policy optimization with behavioral cloning in a feedback loop that recycles verified successful executions into training demonstrations. An object-centered per-hand reward maintains object-motion feedback while supervising each hand according to its reference contact role. Clip-Centered Advantages remove reference-dependent advantage offsets while preserving within-trajectory differences. Together, these designs enable one single policy to learn coordinated manipulation across heterogeneous references. Without fine-tuning, a single checkpoint achieves $86.79\%$ success on unseen trajectories of training objects and $57.32\%$ on held-out objects, surpassing the state-of-the-art bimanual tracking baselines by $77.36$ and $45.73$ percentage points.

\end{abstract}

\section{Introduction}
\label{sec:intro}

Dexterous manipulation supports household assistance, industrial assembly, and assistive care. Unlike gripper-based manipulation, it requires precise coordination of finger motions and changing hand--object contacts to grasp, reorient, and reposition objects~\citep{rajeswaran2018dapg,andrychowicz2020dexterous,chen2023visualdexterity,wan2023unidexgrasppp}. Generalizing across objects and behaviors is challenging, especially when human demonstrations must be retargeted to robot hands with different kinematics~\citep{qin2022dexmv,liu2025dextrack,xu2025dexplore}. Simulation-ready objects~\citep{physx,physxanything,physxomni} and synthetic trajectories~\citep{jing2025humanoidgen,yin2025dexteritygen} enable large-scale interaction data, but may not fully capture natural human manipulation. Recent trackers therefore learn from human hand--object motion~\citep{liu2025dextrack,xu2025dexplore,adalibieke2026adadextrack}, yet remain largely limited to single-hand tasks. Many everyday tasks instead require asymmetric bimanual coordination: one hand stabilizes an object while the other opens, rotates, or repositions it~\citep{fan2023arctic,zhan2024oakink2,zhou2026bidexhd}. This motivates a generalist bimanual reference tracker that learns from diverse human motion to control different objects and tasks with one policy.

Developing such a tracker poses two challenges. First, public datasets include bimanual manipulation~\citep{fan2023arctic,zhan2024oakink2,liu2024taco,fu2025gigahands}, but fully tracked hand--object sequences with sustained two-hand contact remain scarce, limiting the available coordination patterns. Second, a shared policy must learn trajectories with changing hand roles and contact timing. Hand-pose accuracy alone does not ensure successful object manipulation, while requiring both hands to satisfy contact conditions can mask one hand's progress. Unequal clip difficulty also creates advantage offsets that can weaken updates for useful actions in harder sequences. Shared training therefore needs feedback grounded in object motion and each hand's reference contacts, alongside advantages that preserve differences among transitions within each clip.

We address these challenges with \method, a generalist bimanual reference tracker, and BiHOI, a multi-view hand--object interaction dataset. BiHOI contains $1{,}575$ sequences spanning $229$ objects and $14.77$ hours, captured from $32$ calibrated RGB views. It covers symmetric manipulation and asymmetric coordination. We reconstruct both hands and the object, then retarget hand motion to robot references paired with the object trajectory. Combined with public references, BiHOI broadens the training objects and coordination patterns. \method\ learns one policy for both hands through proximal policy optimization (PPO)~\citep{schulman2017ppo} and behavioral cloning (BC) on verified demonstrations. Trajectory-specific policies optimize screened references, while successful shared-policy executions are verified and reused. At inference, one checkpoint tracks new references without further training. Our \emph{object-centered per-hand reward} maintains object-motion feedback and supervises each hand's reference contacts, including asynchronous engagement. Our \emph{Clip-Centered Advantages (CCA)} subtract each clip's mean estimated advantage before PPO updates, removing clip-dependent offsets while preserving within-clip differences. On held-out trajectories and objects from public datasets, \method\ exceeds the stronger bimanual baseline by $77.36$ and $45.73$ percentage points in success rate, respectively.

Our contributions are:
\begin{itemize}[leftmargin=1.5em,itemsep=1pt,topsep=2pt]
\item To our knowledge, \method\ is the first bimanual reference tracker jointly trained across operation categories and human motion datasets; a single checkpoint is evaluated on unseen trajectories and objects without fine-tuning.
\item We introduce BiHOI, a multi-view hand-object interaction dataset emphasizing coordinated bimanual manipulation, and reconstruct its recordings into robot motion references.
\item We propose an object-centered per-hand reward and CCA to support joint PPO training across bimanual references with different contact roles and difficulty.
\item On our two public-data evaluation splits, \method\ outperforms bimanual adaptations of DexTrack and Dexplore in success rate and completion, reaching $86.79\%$ and $57.32\%$ success on unseen trajectories and objects, respectively.
\end{itemize}

\section{Related Work}
\label{sec:related}

\paragraph{From human references to executable manipulation.}
Human trajectories provide motion targets, but retargeting alone cannot ensure successful contact. Sequence-specific policies commonly use reinforcement learning to correct retargeted motion. ManipTrans~\citep{li2025maniptrans} reuses a general hand-motion imitator but learns a task-specific residual policy for interaction. DexMachina~\citep{mandi2025dexmachina} learns from individual demonstrations by gradually removing virtual object assistance. CHORD~\citep{zhu2026chord} guides sequence-specific learning through contact-wrench matching, while ConTrack~\citep{liang2026contrack} trains one policy per clip with an adaptive balance between object tracking and motion fidelity. We instead learn one generalist bimanual policy that tracks diverse references without per-reference training or fine-tuning at test time.

\paragraph{Generalist manipulation policies.}
Generalist policies share knowledge across objects, motions, or skills, predominantly in single-hand settings. UniDexGrasp++~\citep{wan2023unidexgrasppp} generalizes grasping across object geometries, UniCross~\citep{zhang2026unicross} unifies multiple manipulation skills, and ArtManip~\citep{yang2026artmanip} learns category-level articulated in-hand manipulation. For reference tracking, DexTrack~\citep{liu2025dextrack} combines reinforcement learning with demonstrations mined by trajectory-specific controllers. Dexplore~\citep{xu2025dexplore} expands exploration around reference motions, while AdaDexTrack~\citep{adalibieke2026adadextrack} combines specialist distillation with online reference modulation. Bimanual tracking must coordinate complementary hand roles and contact timing. Existing work explores narrower generalization: BiDexHD~\citep{zhou2026bidexhd} distills teachers into a separate trajectory-conditioned policy for each operation category and evaluates new object--tool combinations and instances within categories; PhysGraph~\citep{li2026physgraph} transfers task-trained policies to new tool and object geometries. We jointly train one bimanual reference tracker across operation categories and human motion datasets, and evaluate it on unseen trajectories and objects without fine-tuning.

\paragraph{Bimanual hand--object interaction datasets.}
GRAB~\citep{taheri2020grab} records whole-body grasping with detailed contact information, while ARCTIC~\citep{fan2023arctic} captures the coupled motion of two hands and articulated objects. OakInk2~\citep{zhan2024oakink2} organizes bimanual manipulation into primitive and complex tasks, and TACO~\citep{liu2024taco} covers diverse tool--action--object compositions. GigaHands~\citep{fu2025gigahands} further expands the scale and diversity of bimanual activities with language annotations. Complementing these motion collections, H2O~\citep{kwon2021h2o} and HOI4D~\citep{liu2022hoi4d} support egocentric interaction understanding, while HOT3D~\citep{banerjee2025hot3d} and HO-Cap~\citep{wang2024hocap} provide recordings and annotations for hand--object reconstruction and tracking. Capture and annotation conventions differ, and human kinematics do not ensure executable robot actions. We harmonize, retarget, and screen zero-residual feasibility to form a common bimanual reference pool for joint training.

\section{Bimanual Hand--Object Dataset}
\label{sec:data}

\begin{table}[!ht]
\caption{Dataset overview: source statistics and simultaneous-contact rates on analyzed recordings.}
\label{tab:data}
\centering
\footnotesize
\setlength{\tabcolsep}{2pt}
\renewcommand{\arraystretch}{1.04}
\begin{tabular*}{\linewidth}{@{}l@{\extracolsep{\fill}}lrrrc@{}}
\toprule
Source & Capture & Sequences & Frames & Objects & \shortstack{Bimanual\\contact (\%)} \\
\midrule
\textbf{BiHOI (ours)} & MV RGB & 1{,}575 & $\approx$1.60M$^{\dagger}$ & 229 & \textbf{65.91} \\
\cmidrule(lr){1-6}
GRAB~\citeyearpar{taheri2020grab} & Marker MoCap & 1{,}334 & 1.622M & 51 & 32.21 \\
OakInk2~\citeyearpar{zhan2024oakink2} & MV RGB + MoCap & 627 & 4.01M & 75 & 24.97 \\
TACO (V1)~\citeyearpar{liu2024taco} & MV RGB + ego RGB-D & 2.5K & 5.2M & 206 & 15.69 \\
GigaHands~\citeyearpar{fu2025gigahands} & MV RGB & -- & 183M & 417 & 34.74 \\
H2O~\citeyearpar{kwon2021h2o} & MV RGB-D & -- & 571.6K & 8 & 38.57 \\
HOI4D~\citeyearpar{liu2022hoi4d} & Ego RGB-D & 4{,}000 & 2.4M & 800 & -- \\
HO-Cap (rev.)~\citeyearpar{wang2024hocap} & MV RGB-D & 64 & 656K & 64 & -- \\
\bottomrule
\end{tabular*}
\par\smallskip
\begin{minipage}{\linewidth}
\scriptsize
MV: multi-view; ego: egocentric. $^{\dagger}$14.77 hours at 30\,fps, counting synchronized time steps once. Public frame counts follow source conventions. Contact is the total number of both-hand-contact frames divided by frames with at least one hand in contact, pooled across analyzed clips; --: unavailable.
\end{minipage}
\end{table}

To expand limited bimanual training data, we introduce \textbf{BiHOI}, a multi-view hand--object interaction dataset. We describe its capture system and composition, then the annotation and retargeting pipeline that converts recordings into robot references.

\subsection{Multi-View Capture and Dataset}
\label{sec:data_capture}

\paragraph{Multi-view capture system.}
We record hand--object interactions with $32$ calibrated RGB cameras surrounding a tabletop workspace, at a resolution of $1920\times1232$ pixels and a frame rate of $30$\,fps. Complementary views support reconstruction under occlusion; calibration supplies a common metric coordinate system. 

\paragraph{Dataset composition.}
BiHOI contains $1{,}575$ interaction sequences covering $229$ objects and $14.77$ hours of unique interaction. The objects span household items, containers, utensils, tools, and electronics. Participants perform natural object-centric interactions rather than following fixed motion templates. The recordings include symmetric bimanual manipulation, asymmetric bimanual coordination, and coordinated multi-object manipulation. 
Table~\ref{tab:data} compares BiHOI with commonly used hand--object interaction datasets.
Our collection emphasizes sustained two-hand interaction. Among frames with at least one hand in contact, $65.91\%$ have both hands in contact, exceeding H2O, the highest analyzed public subset at $38.57\%$, by $27.34$ percentage points. 

\subsection{Annotation and Reference Construction}
\label{sec:data_pipeline}

\paragraph{Reconstruction and annotation.}
We reconstruct metric hand--object trajectories from synchronized RGB recordings.
SAM3 provides object masks~\citep{carion2025sam3}, while multi-view Depth Anything 3 provides spatially consistent depth that we align to metric scale using the calibrated camera extrinsics~\citep{lin2025depthanything3}.
MV-SAM3D reconstructs a canonical object mesh from first-frame views and masks~\citep{li2026mvsam3d}; back-projected depth determines its physical scale.
FoundationPose then recovers 6-DoF object motion~\citep{wen2024foundationpose}, using multi-view pose consensus to handle occlusion and re-registration when consensus fails.

For each participant, triangulated calibration keypoints are used to estimate metric hand scale and calibrate a participant-specific MANO shape.
WiLoR provides per-view hand observations~\citep{potamias2025wilor}; these observations are triangulated, temporally smoothed, and fitted with the MANO hand model~\citep{romero2017mano}, with shape held fixed across the sequence.
Joint refinement against multi-view metric depth combines geometric alignment with reconstruction and temporal regularization, producing bimanual MANO parameters, a metric object mesh, and its 6-DoF trajectory in a common world frame.
Reconstruction and refinement details appear in Appendix~\ref{supp:annotation}.

\paragraph{Retargeting.}
We map the reconstructed human motion to the Wuji robot hands through frame-wise kinematic fitting, following standard retargeting practice~\citep{handa2020dexpilot,qin2023anyteleop,liu2025dextrack}. 
After wrist-frame alignment, we fit hand keypoints to robot links under joint limits, initializing and regularizing each frame against the preceding solution. Smoothing occurs only before retargeting. The resulting wrist and finger references are paired with the object trajectory.

\section{Methodology}
\label{sec:method}

\begin{figure}[t]
\centering
\includegraphics[width=\linewidth]{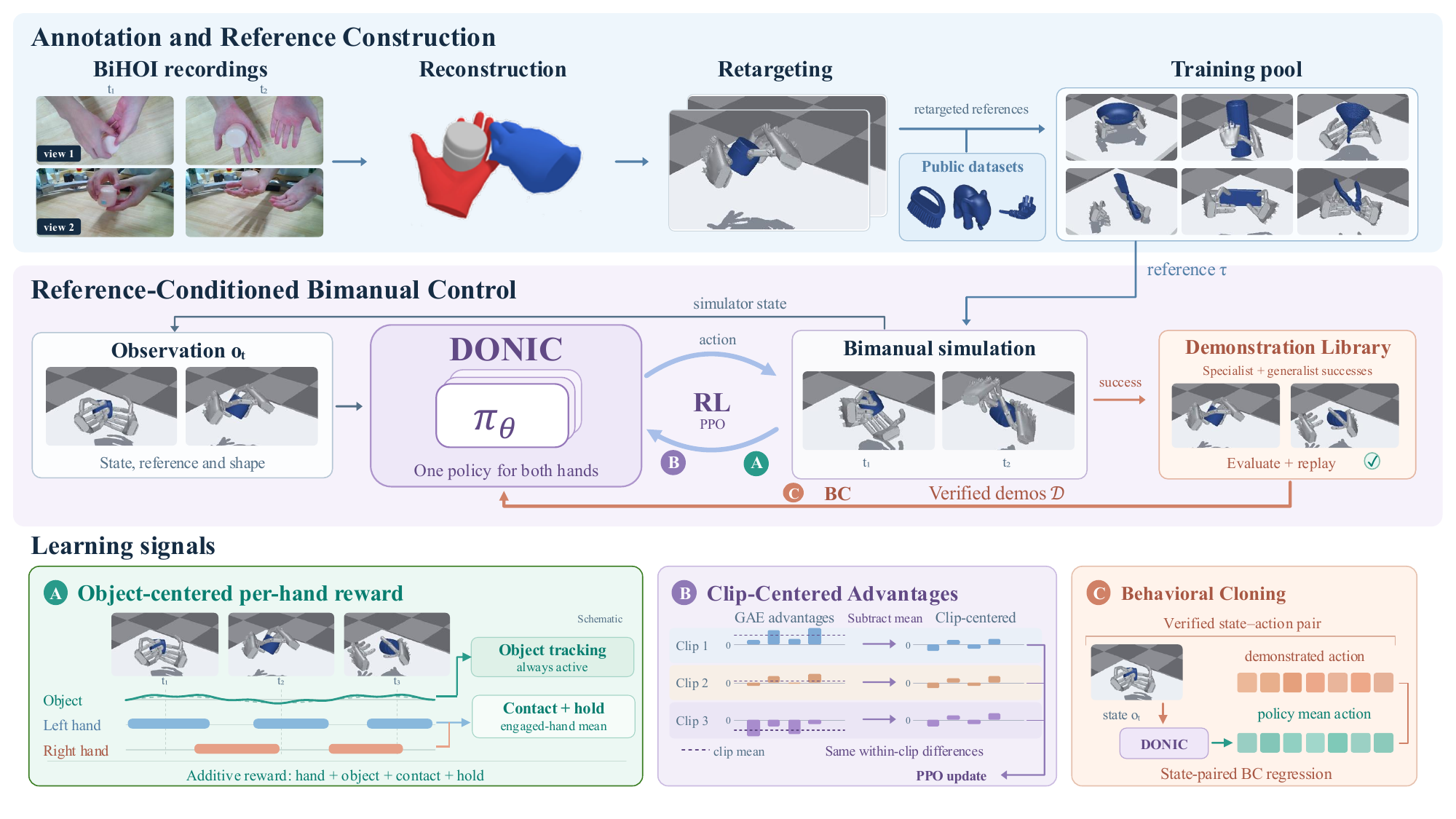}
\caption{\textbf{Overview of \method.}
\textbf{Top:} BiHOI captures are reconstructed and retargeted, then combined with public-data references to form the training pool.
\textbf{Middle:} A shared bimanual policy learns through simulated interaction and verified demonstrations in an iterative PPO+BC loop.
\textbf{Bottom:} An object-centered per-hand reward, Clip-Centered Advantages (CCA), and behavioral cloning provide complementary learning signals.}
\label{fig:overview}
\end{figure}

\method\ learns a generalist bimanual tracker across operation categories and human-data sources (Figure~\ref{fig:overview}). Its object-centered per-hand reward supports complementary roles and asynchronous contacts while maintaining object-motion feedback. CCA removes reference-dependent offsets after generalized advantage estimation (GAE)~\citep{schulman2016gae}, preserving within-reference advantage differences.

\subsection{Task Definition}
\label{sec:setup}

Given a reference for two hands and an object, one policy controls both robot hands to reproduce the object motion through physical interaction. Following DexTrack~\citep{liu2025dextrack}, training combines PPO with BC from executable demonstrations; the policy tracks unseen trajectories and objects without fine-tuning.

For a clip spanning $T$ control steps, we write the reference as
\begin{equation}
\tau=\big\{(\hat{\vq}^{L}_t,\hat{\vq}^{R}_t,\hat{\vp}^{o}_t,\hat{\mR}^{o}_t)\big\}_{t=0}^{T}.
\label{eq:reference}
\end{equation}
Here $t$ is the frame index, $h\in\{L,R\}$ identifies a hand, and a hat denotes a reference quantity. The configuration $\hat{\vq}^{h}_t\in\R^d$ contains the wrist coordinates and finger-joint angles, with $d$ controlled coordinates per hand. The object position and orientation are $\hat{\vp}^{o}_t\in\R^3$ and $\hat{\mR}^{o}_t\in SO(3)$, where $SO(3)$ is the space of three-dimensional rotations. We concatenate the two hand configurations into $\hat{\vq}_t\in\R^{2d}$. Corresponding quantities without hats denote simulated states. Object linear velocity is denoted by $\vv^o_t$; its reference $\hat{\vv}^o_t$ is obtained by finite differences. Positions, rotations, and linear velocities use metres, radians, and metres per second, respectively.

\subsection{Reference-Conditioned Bimanual Control}
\label{sec:learning}

A single stochastic policy $\pi_\theta(\va_t\mid\vo_t)$, parameterized by $\theta$, predicts joint actions $\va_t$ for both hands from observation $\vo_t$. Following reference-based trackers~\citep{liu2025dextrack,xu2025dexplore}, $\vo_t$ concatenates two per-hand blocks containing proprioception, object pose and velocity, upcoming reference targets, reference-state errors, and fingertip-to-object distance vectors. Spatial features use each wrist's heading frame. We append a basis-point-set shape encoding~\citep{prokudin2019bps}, object scale and centroid, hand identities, and the fixed left-hand reward weight $w_L\in(0,1]$. Contact-force channels are zeroed; simulated forces supervise rewards only.

Given $\vo_t$, the policy predicts bounded corrections $\va_t\in[-1,1]^{2d}$ to the next reference, producing a proportional--derivative (PD) control target as in prior trackers~\citep{liu2025dextrack,mandi2025dexmachina}:
\begin{equation}
\vq^{\mathrm{pd}}_t=\operatorname{clamp}\!\left(\hat{\vq}_{t+1}+\vs\odot\va_t,\vq_{\min},\vq_{\max}\right),
\label{eq:action}
\end{equation}
where $\vs\in\R^{2d}$ sets the correction range, $\odot$ denotes element-wise multiplication, and $\operatorname{clamp}$ enforces the coordinate limits $\vq_{\min}$ and $\vq_{\max}$. Corrections are anchored to the next reference at every step and do not accumulate. The PD controller executes the target, and the resulting state supplies the next observation. We build on this control interface with the following reward and training procedure.

\subsection{Object-Centered Per-Hand Reward}
\label{sec:reward}

Hand-pose accuracy does not ensure contact or object motion, and the hands may engage asynchronously with different roles. We therefore reward object motion throughout the clip while supervising each hand's contact only when required by the reference.

\paragraph{Tracking the object.}
We reward agreement in object position, orientation, and linear velocity:
\begin{equation}
 r_t^{\mathrm{obj}}=
 e^{-\kappa_p\|\vp^o_t-\hat{\vp}^o_t\|_2}
 e^{-\kappa_R\angle(\mR^o_t,\hat{\mR}^o_t)}
 e^{-\kappa_v\|\vv^o_t-\hat{\vv}^o_t\|_2}.
\label{eq:robj}
\end{equation}
Positive coefficients $\kappa_p,\kappa_R,\kappa_v$ control sensitivity to position, rotation, and velocity errors. Here $\|\cdot\|_2$ is the Euclidean norm and $\angle(\cdot,\cdot)$ the rotation geodesic distance. The product rewards joint agreement, with velocity penalizing stationary solutions to moving references. This kinematic term~\citep{wang2023physhoi,wang2024skillmimic} is independent of the contact gate.

\paragraph{Supervising each hand's contact.}
Object tracking alone does not specify how the hands should support the motion. Let $\mathcal C_t^h$ be the set of fingertips on hand $h$ required to contact the object by the reference at frame $t$. For fingertip $i$, let $\delta_{i,t}^h$ denote its distance to the object surface and $f_{i,t}^h$ its simulated net contact-force magnitude. We estimate contact using
\begin{equation}
 \alpha_{i,t}^h=\1[f_{i,t}^h>f_{\min}]\,\1[\delta_{i,t}^h<\delta_{\max}],
\label{eq:contact_proxy}
\end{equation}
where $\1[\cdot]$ is an indicator and $f_{\min},\delta_{\max}>0$ are force and distance thresholds. Proximity reduces false object-contact attribution, but net force does not identify the collision partner.

For a hand with at least one required fingertip, we define
\begin{equation}
\begin{aligned}
 r_t^{\mathrm{cont},h}
 &=\frac{1}{|\mathcal C_t^h|}\sum_{i\in\mathcal C_t^h}
 \left[\zeta e^{-\kappa_c\delta_{i,t}^h}+(1-\zeta)\alpha_{i,t}^h\right],\\
 r_t^{\mathrm{hold},h}
 &=\frac{1}{|\mathcal C_t^h|}\sum_{i\in\mathcal C_t^h}\alpha_{i,t}^h.
\end{aligned}
\label{eq:rcont}
\end{equation}
Here $|\mathcal C_t^h|$ is the number of required fingertips, $\kappa_c>0$ controls distance sensitivity, and $\zeta\in[0,1]$ balances proximity and contact. Proximity guides contact acquisition; the hold term rewards instantaneous required-contact coverage, allowing valid sliding or rolling.

\paragraph{Combining the learning signals.}
We aggregate contact rewards over the engaged-hand set $\mathcal E_t=\{h\in\{L,R\}:\mathcal C_t^h\ne\varnothing\}$. For $u\in\{\mathrm{cont},\mathrm{hold}\}$, let
\begin{equation}
 r_t^u=
 \begin{cases}
 \displaystyle\frac{\sum_{h\in\mathcal E_t}w_h r_t^{u,h}}
 {\sum_{h\in\mathcal E_t}w_h},&\mathcal E_t\ne\varnothing,\\[4pt]
 0,&\mathcal E_t=\varnothing,
 \end{cases}
\label{eq:avg}
\end{equation}
where $w_R=1$ and $w_L\in(0,1]$ are fixed hand weights. Averaging preserves either engaged hand's feedback when the other loses contact; excluding inactive hands avoids rewarding attachment during reference release.

Finally, we combine these signals with the inherited hand-tracking reward $r_t^{\mathrm{hand}}\in[0,1]$, which measures wrist, finger-joint, and keypoint agreement~\citep{xu2025dexplore}:
\begin{equation}
 r_t=(1+w_L)\left[
 \beta_{\mathrm{hand}}r_t^{\mathrm{hand}}
 +\beta_{\mathrm{obj}}r_t^{\mathrm{obj}}
 +\beta_{\mathrm{cont}}r_t^{\mathrm{cont}}
 +\beta_{\mathrm{hold}}r_t^{\mathrm{hold}}\right].
\label{eq:reward}
\end{equation}
The nonnegative $\beta$ coefficients sum to one; $(1+w_L)$ sets the reward scale. These terms jointly supervise hand motion, object outcomes, and reference contact roles (settings in Appendix~\ref{supp:reward}).

\subsection{Learning a Generalist from Verified Demonstrations}
\label{sec:optimization}

Following DexTrack~\citep{liu2025dextrack}, we alternate PPO+BC training with demonstration collection. PPO learns through simulated interaction, while BC supplies successful actions that shared exploration may struggle to discover. Trajectory-specific policies optimize references retained by replay screening; their verified rollouts form a demonstration library $\mathcal D$ of observation--action pairs. As the generalist improves, its verified successes are added to $\mathcal D$ for subsequent updates. All demonstrations come from training references.

Demonstrations supply successful behavior, while CCA addresses unequal reference difficulty in PPO updates.

\paragraph{Clip-Centered Advantages.}
Shared PPO batches mix contact patterns and coordination demands. GAE combines temporally weighted value-prediction residuals, but unequal critic accuracy across references can leave clip-dependent offsets. Difficult clips may then receive consistently negative advantages despite relatively successful transitions. CCA removes this common offset while preserving within-clip differences.

Formally, let $\mathcal B$ denote an on-policy rollout batch, $j$ index a transition, and $m(j)$ identify its reference clip. After computing the standard GAE advantage $A_j^{\mathrm{GAE}}$ from rewards and the learned value function, we group transitions by reference as $\mathcal G_m=\{j\in\mathcal B:m(j)=m\}$ and use
\begin{equation}
 \widetilde A_j=A_j^{\mathrm{GAE}}
 -\frac{1}{|\mathcal G_{m(j)}|}
 \sum_{k\in\mathcal G_{m(j)}}A_k^{\mathrm{GAE}}.
\label{eq:groupadv}
\end{equation}
Subtracting the clip mean preserves pairwise advantage differences among its transitions while removing their common empirical offset. The adjustment is applied after GAE; the learned critic and temporal advantage estimation remain unchanged.

Groups are defined by reference identity rather than by environment instance or temporal phase, allowing parallel rollouts of the same reference to contribute to a common baseline. Unlike batch-wide centering, CCA removes an offset specific to each reference instead of applying one shared shift to all clips. We subtract only the group mean, without dividing by a per-clip standard deviation, preserving the scale of within-clip advantage differences. This adjustment acts on the policy's learning signal; it does not modify the reward or the critic's regression targets.

\paragraph{Joint PPO and BC updates.}
We use $\widetilde A_j$ in the standard PPO actor loss $\mathcal L_{\mathrm{PPO}}$, retaining its on-policy probability ratio and clipping rule. For a demonstration observation--action pair $(\vo^D,\va^D)\in\mathcal D$, let $\mu_\theta(\vo^D)$ denote the shared Gaussian policy's mean action. The combined actor objective is
\begin{equation}
 \mathcal L_{\mathrm{actor}}(\theta)=
 \mathcal L_{\mathrm{PPO}}(\theta;\widetilde A)
 +\lambda\E_{(\vo^D,\va^D)\sim\mathcal D}
 \left\|\mu_\theta(\vo^D)-\va^D\right\|_2^2,
\label{eq:bc}
\end{equation}
where $\lambda>0$ weights BC and the expectation averages over sampled demonstration pairs. Demonstrations use a separate sampling budget and do not enter the PPO ratio or GAE. The critic retains its standard value-regression objective. Sampling, verification, and training details are provided in Appendices~\ref{supp:demos}--\ref{supp:training}.

\section{Experiments}
\label{sec:experiments}

\begin{figure}[t]
\centering
\includegraphics[width=\linewidth]{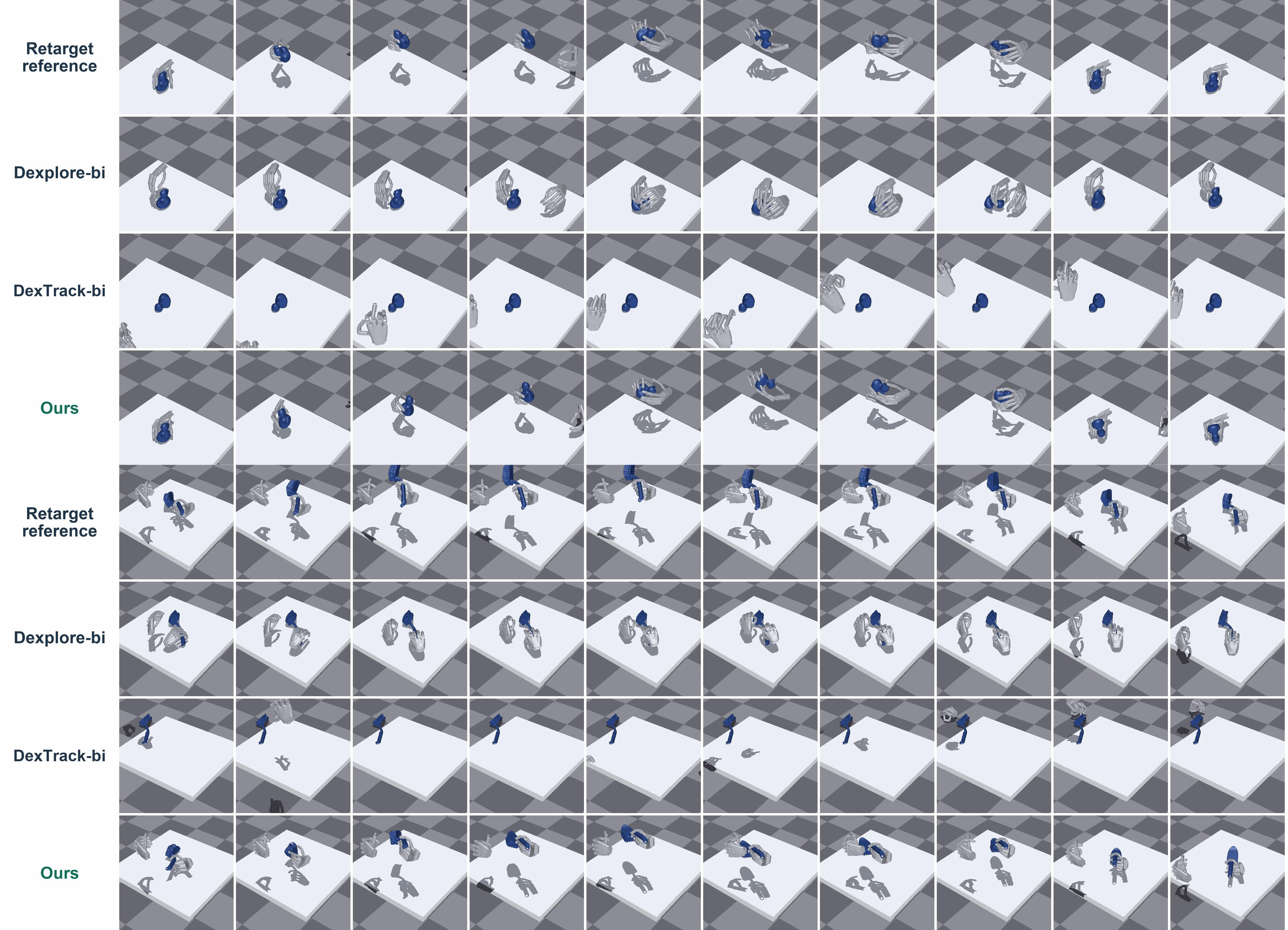}
\caption{Qualitative comparisons on two held-out object trajectories. Rows show the reference, Dexplore-bi, DexTrack-bi, and \method; time runs left to right.}
\label{fig:policy_comparison}
\end{figure}

\subsection{Setup}
\label{sec:exp_setup}

\paragraph{Datasets.}
\label{sec:data_public}
We curate bimanual manipulation references from GRAB~\citep{taheri2020grab}, GigaHands~\citep{fu2025gigahands}, OakInk2~\citep{zhan2024oakink2}, TACO~\citep{liu2024taco}, and H2O~\citep{kwon2021h2o}.
We retain clips in which both hands and the manipulated object are tracked, and each hand contacts the object at some point during the clip; simultaneous contact is not required.
Motions are harmonized, retargeted to the Wuji Hand V1, and replay-screened (Appendix~\ref{supp:inventory}).
This yields $1{,}845$ public-data training clips, separate from the held-out evaluation references. Ours+BiHOI additionally uses all $1{,}575$ BiHOI sequences, bringing its training pool to $3{,}420$ clips.

For evaluation, we reserve two splits from the public datasets:
\textsc{unseen-traj} contains $53$ unseen trajectories of training objects, and \textsc{unseen-obj} contains $164$ clips involving held-out objects.
Neither split is used for policy training or demonstration collection.
All methods are evaluated on the same splits without fine-tuning.

\paragraph{Metrics.}
We extend AdaDexTrack's evaluation protocol~\citep{adalibieke2026adadextrack} to bimanual tracking.
We report object errors ($T_o$, $R_o$), wrist errors ($T_h$, $R_h$), and finger-joint error ($E_{\mathrm{finger}}$), with translation in centimetres and angles in radians.
Wrist errors use robot forward kinematics against retargeted references; finger error averages over 20 finger joints.
Reported hand errors take the worse of the two trajectory-averaged hand errors, then average across trajectories.
SR requires both hands independently to satisfy
$\tfrac12(T_o+T_h)\le10$\,cm,
$\tfrac12(R_o+R_h)\le\pi/3$, and
$E_{\mathrm{finger}}\le2\pi/9$
on trajectory-averaged errors.
Comp is the fraction of frames satisfying all three conditions for both hands, averaged over trajectories.

\paragraph{Baselines.}
We compare against \emph{DexTrack-bi} and \emph{Dexplore-bi}, using their released code with observation and control interfaces adapted to two Wuji hands.
Both baselines share \method's public-data pool, Isaac Gym simulator, robot model, and held-out splits.
DexTrack-bi follows the released specialist--distillation pipeline~\citep{liu2025dextrack}.
Dexplore-bi retains Dexplore's original reward and reference-scoped exploration~\citep{xu2025dexplore}.
Additional implementation details are provided in Appendix~\ref{supp:baselines}.

\subsection{Main Results}
\label{sec:exp_main}

\begin{table}[t]
\caption{Reference tracking without fine-tuning, using one checkpoint per method across both splits. Black and blue bold mark the best and second-best results within each split; ties share highlighting.}
\label{tab:main}
\centering
\small
\setlength{\tabcolsep}{2.5pt}
\renewcommand{\arraystretch}{1.22}
\begin{tabular*}{\linewidth}{@{}l@{\extracolsep{\fill}}lrrrrrrr@{}}
\toprule
Dataset & Method & $T_o\downarrow$ & $R_o\downarrow$ & $T_h\downarrow$ & $R_h\downarrow$ & $E_{\mathrm{finger}}\downarrow$ & SR$\uparrow$ & Comp$\uparrow$ \\
 & & (cm) & (rad) & (cm) & (rad) & (rad) & (\%) & (\%) \\
\midrule
\multirow{6}{*}{\shortstack[l]{Unseen\\trajectory}}
 & DexTrack-bi & 41.86 & 2.3340 & 23.03 & 0.8296 & 0.6301 & 0.00 & 1.30 \\
 & Dexplore-bi & 14.73 & \textbf{0.8597} & 21.88 & 0.8412 & 0.3166 & 9.43 & 27.87 \\
\cmidrule(l){2-9}
 & w/o CCA & 11.27 & 0.9585 & 4.34 & 0.2980 & 0.0433 & 81.13 & 62.72 \\
 & w/o contact gate & \textcolor{blue}{\textbf{10.26}} & 0.9589 & 4.00 & 0.3013 & \textbf{0.0338} & 81.13 & 65.03 \\
\cmidrule(l){2-9}
 & Ours & 10.27 & 0.9277 & \textcolor{blue}{\textbf{3.96}} & \textcolor{blue}{\textbf{0.2905}} & \textcolor{blue}{\textbf{0.0372}} & \textcolor{blue}{\textbf{86.79}} & \textcolor{blue}{\textbf{67.62}} \\
 & Ours+BiHOI & \textbf{9.74} & \textcolor{blue}{\textbf{0.8637}} & \textbf{3.32} & \textbf{0.2818} & \textbf{0.0338} & \textbf{88.68} & \textbf{68.72} \\
\midrule
\multirow{6}{*}{\shortstack[l]{Unseen\\object}}
 & DexTrack-bi & 127.87 & 2.3532 & 72.83 & 1.0629 & 0.6455 & 0.00 & 0.88 \\
 & Dexplore-bi & 17.92 & 1.1431 & 19.46 & 0.9509 & 0.3685 & 11.59 & 26.37 \\
\cmidrule(l){2-9}
 & w/o CCA & 16.83 & 1.1464 & 5.52 & 0.3235 & 0.0378 & 48.17 & 43.89 \\
 & w/o contact gate & 17.61 & 1.2041 & 4.97 & 0.3386 & \textcolor{blue}{\textbf{0.0371}} & 48.78 & 49.52 \\
\cmidrule(l){2-9}
 & Ours & \textcolor{blue}{\textbf{15.74}} & \textcolor{blue}{\textbf{1.1310}} & \textcolor{blue}{\textbf{4.58}} & \textcolor{blue}{\textbf{0.3226}} & 0.0373 & \textcolor{blue}{\textbf{57.32}} & \textcolor{blue}{\textbf{54.69}} \\
 & Ours+BiHOI & \textbf{13.96} & \textbf{1.1157} & \textbf{4.54} & \textbf{0.3206} & \textbf{0.0361} & \textbf{59.76} & \textbf{55.35} \\
\bottomrule
\end{tabular*}
\end{table}

One \method\ checkpoint improves both held-out splits without fine-tuning (Table~\ref{tab:main}).
On \textsc{unseen-traj}, \method\ achieves $86.79\%$ SR and $67.62\%$ Comp, exceeding the stronger baseline, Dexplore-bi, by $77.36$ and $39.75$ percentage points, respectively.
On \textsc{unseen-obj}, it reaches $57.32\%$ SR and $54.69\%$ Comp, compared with $11.59\%$ and $26.37\%$ for Dexplore-bi.
The gains extend to hand tracking: wrist translation error decreases from $21.88$ to $3.96$\,cm on unseen trajectories and from $19.46$ to $4.58$\,cm on unseen objects.
\method\ reduces all reported errors relative to both baselines except object rotation on \textsc{unseen-traj}.
Because SR and Comp require both hands to meet the tracking criteria, these improvements reflect successful bimanual execution rather than accuracy from only one hand.

On the two held-out object trajectories in Figure~\ref{fig:policy_comparison}, \method\ coordinates both hands and more closely reproduces reference object motion.
Dexplore-bi moves its hands but produces limited object motion, whereas DexTrack-bi loses the hand--object alignment needed to execute the reference.

\subsection{Ablation Studies}
\label{sec:ablation}

\paragraph{Learning signals.}
Removing CCA reduces SR from $86.79\%$ to $81.13\%$ on unseen trajectories and from $57.32\%$ to $48.17\%$ on unseen objects, while lowering Comp and increasing all five tracking errors (Table~\ref{tab:main}). Removing the contact-ownership gate reduces SR to $81.13\%$ and $48.78\%$, respectively. Despite lower individual errors in some cases, this variant also has lower Comp on both splits. Both learning signals thus improve joint bimanual success and completion.

\paragraph{Additional bimanual data.}
Adding all $1{,}575$ BiHOI sequences expands training from $1{,}845$ to $3{,}420$ clips and improves all seven metrics on both splits (Ours+BiHOI, Table~\ref{tab:main}). SR reaches $88.68\%$ and $59.76\%$ on unseen trajectories and objects, with Comp of $68.72\%$ and $55.35\%$. Unseen-object translation error falls by $11.3\%$, from $15.74$ to $13.96$\,cm. Ranking first on all unseen-object metrics, Ours+BiHOI demonstrates the benefit of additional bimanual data.

\section{Conclusion, Limitations and Future Work}
We presented \method, a generalist bimanual tracker, and BiHOI, a multi-view bimanual interaction dataset. Object-centered per-hand rewards and CCA support iterative PPO+BC training, yielding one checkpoint that generalizes to unseen trajectories and objects without fine-tuning and substantially outperforms adapted baselines. Ablations support both learning signals and the benefit of additional BiHOI data.
However, our quantitative evaluation focuses on simulated single-object interactions and relies on reconstructed, retargeted motion references.
We aim to extend \method\ to longer-horizon and multi-object manipulation, evaluate on an independent benchmark, and investigate transfer to physical bimanual systems.

{\small
\bibliography{references}

@inproceedings{liu2025dextrack,
  title     = {{DexTrack}: Towards Generalizable Neural Tracking Control for Dexterous Manipulation from Human References},
  author    = {Liu, Xueyi and Adalibieke, Jianibieke and Han, Qianwei and Qin, Yuzhe and Yi, Li},
  booktitle = {International Conference on Learning Representations (ICLR)},
  year      = {2025},
  url = {https://arxiv.org/abs/2502.09614}
}

@inproceedings{xu2025dexplore,
  title     = {Dexplore: Scalable Neural Control for Dexterous Manipulation from Reference-Scoped Exploration},
  author    = {Xu, Sirui and Chao, Yu-Wei and Bian, Liuyu and Mousavian, Arsalan and Wang, Yu-Xiong and Gui, Liang-Yan and Yang, Wei},
  booktitle = {Conference on Robot Learning (CoRL)},
  year      = {2025},
  url = {https://arxiv.org/abs/2509.09671}
}

@inproceedings{adalibieke2026adadextrack,
  title     = {{AdaDexTrack}: Dynamic Modulation for Adaptive and Generalizable Dexterous Manipulation Tracking},
  author    = {Adalibieke, Jianibieke and Han, Qianwei and Liu, Xueyi and Qin, Yuzhe and Yi, Li},
  booktitle = {IEEE/CVF Conference on Computer Vision and Pattern Recognition (CVPR)},
  year      = {2026},
  url = {https://janebek.github.io/AdaDexTrack/}
}

@inproceedings{li2025maniptrans,
  title     = {{ManipTrans}: Efficient Dexterous Bimanual Manipulation Transfer via Residual Learning},
  author    = {Li, Kailin and Li, Puhao and Liu, Tengyu and Li, Yuyang and Huang, Siyuan},
  booktitle = {IEEE/CVF Conference on Computer Vision and Pattern Recognition (CVPR)},
  year      = {2025},
  url = {https://arxiv.org/abs/2503.21860}
}

@inproceedings{mandi2025dexmachina,
  title     = {{DexMachina}: Functional Retargeting for Bimanual Dexterous Manipulation},
  author    = {Mandi, Zhao and Hou, Yifan and Fox, Dieter and Narang, Yashraj and Mandlekar, Ajay and Song, Shuran},
  booktitle = {International Conference on Machine Learning (ICML)},
  year      = {2026},
  url = {https://project-dexmachina.github.io/}
}

@article{zhu2026chord,
  title   = {Learning Dexterous Manipulation Using Contact Wrench Guidance From Human Demonstration},
  author  = {Zhu, Xinghao and Liu, Zixi and Jain, Shalin and Li, Chenran and Noori, Milad and Lin, Michael Andres and Zhao, Huihua and Welsh, John and Verghese, Mrinal and Liu, Wei and Wang, Tingwu and Da, Xingye and Luo, Zhengyi and Kulkarni, Vishal and Bhatti, Naema and Zhu, Yuke and Fan, Linxi and Wen, Bowen and Xu, Danfei and Pouya, Soha and Chang, Yan},
  journal = {arXiv preprint arXiv:2607.00033},
  year    = {2026},
  url = {https://arxiv.org/abs/2607.00033}
}

@article{liang2026contrack,
  title   = {{ConTrack}: Constrained Hand Motion Tracking with Adaptive Trade-off Control},
  author  = {Liang, Yutong and Peng, Quanquan and Qiu, Ri-Zhao and Wang, Xiaolong},
  journal = {arXiv preprint arXiv:2606.03177},
  year    = {2026},
  url = {https://arxiv.org/abs/2606.03177}
}

@article{zhang2026unicross,
  title   = {{UniCross}: Unified Cross-Skill Dexterous Manipulation Synthesis},
  author  = {Zhang, Hui and Ferchow, Julian and Song, Jie and Meboldt, Mirko},
  journal = {arXiv preprint arXiv:2607.28198},
  year    = {2026},
  url = {https://arxiv.org/abs/2607.28198}
}

@article{wang2023physhoi,
  title   = {{PhysHOI}: Physics-Based Imitation of Dynamic Human-Object Interaction},
  author  = {Wang, Yinhuai and Lin, Jing and Zeng, Ailing and Luo, Zhengyi and Zhang, Jian and Zhang, Lei},
  journal = {arXiv preprint arXiv:2312.04393},
  year    = {2023}
}

@inproceedings{zhou2026bidexhd,
  title     = {Learning Diverse Bimanual Dexterous Manipulation Skills from Human Demonstrations},
  author    = {Zhou, Bohan and Yuan, Haoqi and Fu, Yuhui and Lu, Zongqing},
  booktitle = {Proceedings of the AAAI Conference on Artificial Intelligence},
  year      = {2026},
  url = {https://doi.org/10.1609/aaai.v40i34.40127}
}

@inproceedings{taheri2020grab,
  title     = {{GRAB}: A Dataset of Whole-Body Human Grasping of Objects},
  author    = {Taheri, Omid and Ghorbani, Nima and Black, Michael J. and Tzionas, Dimitrios},
  booktitle = {European Conference on Computer Vision (ECCV)},
  year      = {2020}
}

@inproceedings{fu2025gigahands,
  title     = {{GigaHands}: A Massive Annotated Dataset of Bimanual Hand Activities},
  author    = {Fu, Rao and Zhang, Dingxi and Jiang, Alex and Fu, Wanjia and Funk, Austin and Ritchie, Daniel and Sridhar, Srinath},
  booktitle = {IEEE/CVF Conference on Computer Vision and Pattern Recognition (CVPR)},
  year      = {2025},
  pages     = {17461--17474},
  url       = {https://openaccess.thecvf.com/content/CVPR2025/html/Fu_GigaHands_A_Massive_Annotated_Dataset_of_Bimanual_Hand_Activities_CVPR_2025_paper.html}
}

@inproceedings{zhan2024oakink2,
  title     = {{OakInk2}: A Dataset of Bimanual Hands-Object Manipulation in Complex Task Completion},
  author    = {Zhan, Xinyu and Yang, Lixin and Zhao, Yifei and Mao, Kangrui and Xu, Hanlin and Lin, Zenan and Li, Kailin and Lu, Cewu},
  booktitle = {IEEE/CVF Conference on Computer Vision and Pattern Recognition (CVPR)},
  year      = {2024},
  pages     = {445--456},
  url       = {https://openaccess.thecvf.com/content/CVPR2024/html/Zhan_OAKINK2_A_Dataset_of_Bimanual_Hands-Object_Manipulation_in_Complex_Task_CVPR_2024_paper.html}
}

@inproceedings{liu2024taco,
  title     = {{TACO}: Benchmarking Generalizable Bimanual Tool-{AC}tion-{O}bject Understanding},
  author    = {Liu, Yun and Yang, Haolin and Si, Xu and Liu, Ling and Li, Zipeng and Zhang, Yuxiang and Liu, Yebin and Yi, Li},
  booktitle = {IEEE/CVF Conference on Computer Vision and Pattern Recognition (CVPR)},
  year      = {2024},
  pages     = {21740--21751},
  url       = {https://openaccess.thecvf.com/content/CVPR2024/html/Liu_TACO_Benchmarking_Generalizable_Bimanual_Tool-ACtion-Object_Understanding_CVPR_2024_paper.html}
}

@inproceedings{kwon2021h2o,
  title     = {{H2O}: Two Hands Manipulating Objects for First Person Interaction Recognition},
  author    = {Kwon, Taein and Tekin, Bugra and St{\"u}hmer, Jan and Bogo, Federica and Pollefeys, Marc},
  booktitle = {IEEE/CVF International Conference on Computer Vision (ICCV)},
  year      = {2021}
}

@inproceedings{liu2022hoi4d,
  title     = {{HOI4D}: A {4D} Egocentric Dataset for Category-Level Human-Object Interaction},
  author    = {Liu, Yunze and Liu, Yun and Jiang, Che and Lyu, Kangbo and Wan, Weikang and Shen, Hao and Liang, Boqiang and Fu, Zhoujie and Wang, He and Yi, Li},
  booktitle = {IEEE/CVF Conference on Computer Vision and Pattern Recognition (CVPR)},
  year      = {2022}
}

@inproceedings{wang2024hocap,
  title   = {{HO-Cap}: A Capture System and Dataset for {3D} Reconstruction and Pose Tracking of Hand-Object Interaction},
  author  = {Wang, Jikai and Zhang, Qifan and Chao, Yu-Wei and Wen, Bowen and Guo, Xiaohu and Xiang, Yu},
  booktitle = {Advances in Neural Information Processing Systems},
  volume  = {38},
  year    = {2025},
  doi     = {10.52202/085713-5683}
}

@inproceedings{banerjee2025hot3d,
  title     = {{HOT3D}: Hand and Object Tracking in {3D} from Egocentric Multi-View Videos},
  author    = {Banerjee, Prithviraj and Shkodrani, Sindi and Moulon, Pierre and Hampali, Shreyas and Han, Shangchen and Zhang, Fan and Zhang, Linguang and Fountain, Jade and Miller, Edward and Basol, Selen and Newcombe, Richard and Wang, Robert and Engel, Jakob Julian and Hodan, Tomas},
  booktitle = {IEEE/CVF Conference on Computer Vision and Pattern Recognition (CVPR)},
  year      = {2025},
  pages     = {7061--7071},
  url       = {https://openaccess.thecvf.com/content/CVPR2025/html/Banerjee_HOT3D_Hand_and_Object_Tracking_in_3D_from_Egocentric_Multi-View_CVPR_2025_paper.html}
}

@inproceedings{fan2023arctic,
  title     = {{ARCTIC}: A Dataset for Dexterous Bimanual Hand-Object Manipulation},
  author    = {Fan, Zicong and Taheri, Omid and Tzionas, Dimitrios and Kocabas, Muhammed and Kaufmann, Manuel and Black, Michael J. and Hilliges, Otmar},
  booktitle = {IEEE/CVF Conference on Computer Vision and Pattern Recognition (CVPR)},
  year      = {2023},
  pages     = {12943--12954},
  url       = {https://openaccess.thecvf.com/content/CVPR2023/html/Fan_ARCTIC_A_Dataset_for_Dexterous_Bimanual_Hand-Object_Manipulation_CVPR_2023_paper.html}
}

@article{schulman2017ppo,
  title   = {Proximal Policy Optimization Algorithms},
  author  = {Schulman, John and Wolski, Filip and Dhariwal, Prafulla and Radford, Alec and Klimov, Oleg},
  journal = {arXiv preprint arXiv:1707.06347},
  year    = {2017},
  url       = {https://arxiv.org/abs/1707.06347}
}

@inproceedings{schulman2016gae,
  title     = {High-Dimensional Continuous Control Using Generalized Advantage Estimation},
  author    = {Schulman, John and Moritz, Philipp and Levine, Sergey and Jordan, Michael and Abbeel, Pieter},
  booktitle = {International Conference on Learning Representations (ICLR)},
  year      = {2016}
}

@inproceedings{makoviychuk2021isaacgym,
  title     = {Isaac {Gym}: High Performance {GPU}-Based Physics Simulation For Robot Learning},
  author    = {Makoviychuk, Viktor and Wawrzyniak, Lukasz and Guo, Yunrong and Lu, Michelle and Storey, Kier and Macklin, Miles and Hoeller, David and Rudin, Nikita and Allshire, Arthur and Handa, Ankur and State, Gavriel},
  booktitle = {Advances in Neural Information Processing Systems (NeurIPS) Datasets and Benchmarks Track},
  year      = {2021}
}

@article{andrychowicz2020dexterous,
  title   = {Learning Dexterous In-Hand Manipulation},
  author  = {Andrychowicz, Marcin and Baker, Bowen and Chociej, Maciek and J{\'o}zefowicz, Rafal and McGrew, Bob and Pachocki, Jakub and Petron, Arthur and Plappert, Matthias and Powell, Glenn and Ray, Alex and Schneider, Jonas and Sidor, Szymon and Tobin, Josh and Welinder, Peter and Weng, Lilian and Zaremba, Wojciech},
  journal = {The International Journal of Robotics Research},
  volume  = {39},
  number  = {1},
  pages   = {3--20},
  year    = {2020},
  url       = {https://arxiv.org/abs/1808.00177}
}

@article{chen2023visualdexterity,
  title   = {Visual Dexterity: In-Hand Reorientation of Novel and Complex Object Shapes},
  author  = {Chen, Tao and Tippur, Megha and Wu, Siyang and Kumar, Vikash and Adelson, Edward and Agrawal, Pulkit},
  journal = {Science Robotics},
  volume  = {8},
  number  = {84},
  year    = {2023},
  url       = {https://arxiv.org/abs/2211.11744}
}

@inproceedings{wan2023unidexgrasppp,
  title     = {{UniDexGrasp++}: Improving Dexterous Grasping Policy Learning via Geometry-aware Curriculum and Iterative Generalist-Specialist Learning},
  author    = {Wan, Weikang and Geng, Haoran and Liu, Yun and Shan, Zikang and Yang, Yaodong and Yi, Li and Wang, He},
  booktitle = {IEEE/CVF International Conference on Computer Vision (ICCV)},
  year      = {2023},
  pages     = {3891--3902},
  url       = {https://openaccess.thecvf.com/content/ICCV2023/html/Wan_UniDexGrasp_Improving_Dexterous_Grasping_Policy_Learning_via_Geometry-Aware_Curriculum_and_ICCV_2023_paper.html}
}

@inproceedings{qin2022dexmv,
  title     = {{DexMV}: Imitation Learning for Dexterous Manipulation from Human Videos},
  author    = {Qin, Yuzhe and Wu, Yueh-Hua and Liu, Shaowei and Jiang, Hanwen and Yang, Ruihan and Fu, Yang and Wang, Xiaolong},
  booktitle = {European Conference on Computer Vision (ECCV)},
  year      = {2022},
  url       = {https://www.ecva.net/papers/eccv_2022/papers_ECCV/papers/136990562.pdf}
}

@inproceedings{handa2020dexpilot,
  title     = {{DexPilot}: Vision-Based Teleoperation of Dexterous Robotic Hand-Arm System},
  author    = {Handa, Ankur and Van Wyk, Karl and Yang, Wei and Liang, Jacky and Chao, Yu-Wei and Wan, Qian and Birchfield, Stan and Ratliff, Nathan and Fox, Dieter},
  booktitle = {IEEE International Conference on Robotics and Automation (ICRA)},
  year      = {2020}
}

@inproceedings{qin2023anyteleop,
  title     = {{AnyTeleop}: A General Vision-Based Dexterous Robot Arm-Hand Teleoperation System},
  author    = {Qin, Yuzhe and Yang, Wei and Huang, Binghao and Van Wyk, Karl and Su, Hao and Wang, Xiaolong and Chao, Yu-Wei and Fox, Dieter},
  booktitle = {Robotics: Science and Systems (RSS)},
  year      = {2023}
}

@inproceedings{rajeswaran2018dapg,
  title     = {Learning Complex Dexterous Manipulation with Deep Reinforcement Learning and Demonstrations},
  author    = {Rajeswaran, Aravind and Kumar, Vikash and Gupta, Abhishek and Vezzani, Giulia and Schulman, John and Todorov, Emanuel and Levine, Sergey},
  booktitle = {Robotics: Science and Systems (RSS)},
  year      = {2018},
  url       = {https://arxiv.org/abs/1709.10087}
}

@inproceedings{prokudin2019bps,
  title     = {Efficient Learning on Point Clouds with Basis Point Sets},
  author    = {Prokudin, Sergey and Lassner, Christoph and Romero, Javier},
  booktitle = {IEEE/CVF International Conference on Computer Vision (ICCV)},
  year      = {2019}
}

@inproceedings{wang2024skillmimic,
  title   = {{SkillMimic}: Learning Basketball Interaction Skills from Demonstrations},
  author  = {Wang, Yinhuai and Zhao, Qihan and Yu, Runyi and Tsui, Hok Wai and Zeng, Ailing and Lin, Jing and Luo, Zhengyi and Yu, Jiwen and Li, Xiu and Chen, Qifeng and Zhang, Jian and Zhang, Lei and Tan, Ping},
  booktitle = {IEEE/CVF Conference on Computer Vision and Pattern Recognition (CVPR)},
  year    = {2025},
  url = {https://arxiv.org/abs/2408.15270}
}

@article{yang2026artmanip,
  title = {{ArtManip: Category-Level Articulated In-Hand Manipulation}},
  author = {Yang, Yang and Liu, Tengyu and Li, Puhao and Chen, Zeyuan and Li, Yuyang and Wang, Xingwan and Wu, Yingying and Cui, Zhaopeng and Huang, Siyuan},
  journal = {arXiv preprint arXiv:2609.12498},
  year = {2026},
  url = {https://arxiv.org/abs/2609.12498}
}

@article{li2026physgraph,
  title = {{PhysGraph: Physically-Grounded Graph-Transformer Policies for Bimanual Dexterous Hand--Tool--Object Manipulation}},
  author = {Li, Runfa Blark and Kim, David and Liu, Xinshuang and Suzuki, Keito and Bhatt, Dwait and Raicevic, Nikola and Lin, Xin and Lee, Ki Myung Brian and Atanasov, Nikolay and Nguyen, Truong},
  journal = {arXiv preprint arXiv:2603.01436},
  year = {2026},
  url = {https://arxiv.org/abs/2603.01436}
}

@inproceedings{carion2025sam3,
  title     = {{SAM 3}: Segment Anything with Concepts},
  author    = {Carion, Nicolas and Gustafson, Laura and Hu, Yuan-Ting and Debnath, Shoubhik and Hu, Ronghang and Suris, Didac and Ryali, Chaitanya and Alwala, Kalyan Vasudev and Khedr, Haitham and Huang, Andrew and Lei, Jie and Ma, Tengyu and Guo, Baishan and Kalla, Arpit and Marks, Markus and Greer, Joseph and Wang, Meng and Sun, Peize and R{\"a}dle, Roman and Afouras, Triantafyllos and Mavroudi, Effrosyni and Xu, Katherine and Wu, Tsung-Han and Zhou, Yu and Momeni, Liliane and Hazra, Rishi and Ding, Shuangrui and Vaze, Sagar and Porcher, Francois and Li, Feng and Li, Siyuan and Kamath, Aishwarya and Cheng, Ho Kei and Doll{\'a}r, Piotr and Ravi, Nikhila and Saenko, Kate and Zhang, Pengchuan and Feichtenhofer, Christoph},
  booktitle = {International Conference on Learning Representations (ICLR)},
  year      = {2026}
}

@inproceedings{lin2025depthanything3,
  title     = {{Depth Anything 3}: Recovering the Visual Space from Any Views},
  author    = {Lin, Haotong and Chen, Sili and Liew, Junhao and Chen, Donny Y. and Li, Zhenyu and Shi, Guang and Feng, Jiashi and Kang, Bingyi},
  booktitle = {International Conference on Learning Representations (ICLR)},
  year      = {2026}
}

@article{li2026mvsam3d,
  title   = {{MV-SAM3D}: Adaptive Multi-View Fusion for Layout-Aware {3D} Generation},
  author  = {Li, Baicheng and Wu, Dong and Li, Jun and Zhou, Shunkai and Zeng, Zecui and Li, Lusong and Zha, Hongbin},
  journal = {arXiv preprint arXiv:2603.11633},
  year    = {2026},
  url     = {https://arxiv.org/abs/2603.11633}
}

@inproceedings{wen2024foundationpose,
  title     = {{FoundationPose}: Unified {6D} Pose Estimation and Tracking of Novel Objects},
  author    = {Wen, Bowen and Yang, Wei and Kautz, Jan and Birchfield, Stan},
  booktitle = {IEEE/CVF Conference on Computer Vision and Pattern Recognition (CVPR)},
  pages     = {17868--17879},
  year      = {2024},
  doi       = {10.1109/CVPR52733.2024.01692}
}

@article{romero2017mano,
  title   = {Embodied Hands: Modeling and Capturing Hands and Bodies Together},
  author  = {Romero, Javier and Tzionas, Dimitrios and Black, Michael J.},
  journal = {ACM Transactions on Graphics},
  volume  = {36},
  number  = {6},
  pages   = {245:1--245:17},
  year    = {2017},
  doi     = {10.1145/3130800.3130883}
}

@inproceedings{potamias2025wilor,
  title     = {{WiLoR}: End-to-End {3D} Hand Localization and Reconstruction in-the-Wild},
  author    = {Potamias, Rolandos Alexandros and Zhang, Jinglei and Deng, Jiankang and Zafeiriou, Stefanos},
  booktitle = {IEEE/CVF Conference on Computer Vision and Pattern Recognition (CVPR)},
  pages     = {12242--12254},
  year      = {2025},
  doi       = {10.1109/CVPR52734.2025.01143}
}

@inproceedings{jing2025humanoidgen,
  title     = {{HumanoidGen}: Data Generation for Bimanual Dexterous Manipulation via {LLM} Reasoning},
  author    = {Jing, Zhi and Yang, Siyuan and Ao, Jicong and Xiao, Ting and Jiang, Yu-Gang and Bai, Chenjia},
  booktitle = {Advances in Neural Information Processing Systems (NeurIPS)},
  volume    = {38},
  pages     = {173413--173459},
  year      = {2025}
}

@inproceedings{physx,
  title     = {{PhysX-3D}: Physical-Grounded {3D} Asset Generation},
  author    = {Cao, Ziang and Chen, Zhaoxi and Pan, Liang and Liu, Ziwei},
  booktitle = {Advances in Neural Information Processing Systems (NeurIPS)},
  volume    = {38},
  pages     = {104070--104083},
  year      = {2025}
}

@inproceedings{yin2025dexteritygen,
  title     = {{DexterityGen}: Foundation Controller for Unprecedented Dexterity},
  author    = {Yin, Zhao-Heng and Wang, Changhao and Pineda, Luis and Hogan, Francois and Bodduluri, Krishna and Sharma, Akash and Lancaster, Patrick and Prasad, Ishita and Kalakrishnan, Mrinal and Malik, Jitendra and Lambeta, Mike and Wu, Tingfan and Abbeel, Pieter and Mukadam, Mustafa},
  booktitle = {Robotics: Science and Systems (RSS)},
  year      = {2025}
}

@inproceedings{physxanything,
  title     = {{PhysX-Anything}: Simulation-Ready Physical {3D} Assets from Single Image},
  author    = {Cao, Ziang and Hong, Fangzhou and Chen, Zhaoxi and Pan, Liang and Liu, Ziwei},
  booktitle = {IEEE/CVF Conference on Computer Vision and Pattern Recognition (CVPR)},
  pages     = {5839--5848},
  year      = {2026}
}

@article{physxomni,
  title   = {{PhysX-Omni}: Unified Simulation-Ready Physical {3D} Generation for Rigid, Deformable, and Articulated Objects},
  author  = {Cao, Ziang and Liu, Yinghao and Li, Haitian and Yao, Runmao and Hong, Fangzhou and Chen, Zhaoxi and Pan, Liang and Liu, Ziwei},
  journal = {arXiv preprint arXiv:2605.21572},
  year    = {2026}
}
}

\newpage
\appendix
\startcontents[appendix]
\printcontents[appendix]{l}{1}{%
  \section*{Appendix Contents}%
  \vspace{-0.5em}%
}
\newpage
This appendix provides additional related work (Appendix~\ref{supp:related}), dataset construction and retargeting details (Appendix~\ref{supp:data}), controller and training specifications (Appendix~\ref{supp:method}), and experimental protocols and results (Appendix~\ref{supp:exp}).

\providecommand{\ind}{\mathbf 1}
\providecommand{\rotangle}{d_{\mathrm{SO}(3)}}
\setcounter{equation}{0}
\setcounter{table}{0}
\setcounter{figure}{0}
\renewcommand{\theequation}{\thesection.\arabic{equation}}
\renewcommand{\thetable}{S\arabic{table}}
\renewcommand{\thefigure}{S\arabic{figure}}
\renewcommand{\theHequation}{supp.\thesection.\arabic{equation}}
\renewcommand{\theHtable}{supp.\arabic{table}}
\renewcommand{\theHfigure}{supp.\arabic{figure}}

\section{Additional Related Work}
\label{supp:related}

\paragraph{Executable human references and shared control.}
Human demonstrations provide motion targets but do not directly specify robot actions: differences in hand kinematics and simulated contact can make a kinematically plausible reference physically difficult to execute. Reference-based trackers address this gap through closed-loop control. DexTrack~\citep{liu2025dextrack} combines trajectory-specific demonstration generation with generalist learning; Dexplore~\citep{xu2025dexplore} expands exploration around reference motions; and AdaDexTrack~\citep{adalibieke2026adadextrack} combines specialist distillation with reference modulation. Our reference-conditioned action interface and iterative reinforcement--imitation learning loop build on this line of work. Our contributions within that loop are the object-centered per-hand reward and Clip-Centered Advantages (CCA), designed for shared bimanual tracking across heterogeneous references.

\paragraph{What generalization means here.}
Generalization across grasped objects, manipulation skills, and complete reference trajectories imposes different requirements. UniDexGrasp++~\citep{wan2023unidexgrasppp} studies generalizable dexterous grasping, while BiDexHD~\citep{zhou2026bidexhd} learns trajectory-conditioned bimanual policies within operation categories. Our controller is jointly trained across operation categories and human-motion sources. We evaluate new trajectories of training objects and trajectories involving held-out objects with the same checkpoint. These two splits test reference and object generalization; they do not, by themselves, establish generalization to held-out operation categories or arbitrary manipulation goals.

\paragraph{Interaction capture and usable control data.}
GRAB, GigaHands, OakInk2, TACO, and H2O provide complementary sources of hand--object interaction~\citep{taheri2020grab,fu2025gigahands,zhan2024oakink2,liu2024taco,kwon2021h2o}. HOI4D and HO-Cap further broaden capture settings and annotation tools~\citep{liu2022hoi4d,wang2024hocap}. Our comparison distinguishes the inventory of a source dataset from the subset retained for robot training. A clip enters our bimanual pool only if the required hand and object tracks are available and both hands contact the object at some point. This allows sequential as well as simultaneous hand participation. BiHOI complements these sources with a higher measured proportion of simultaneous bimanual contact in the analyzed recordings.

\section{Dataset and Reference Construction}
\label{supp:data}
\setcounter{equation}{0}

\subsection{Capture, Composition, and Training Inventory}
\label{supp:inventory}

BiHOI contains $1{,}575$ interaction sequences covering $229$ objects and $14.77$ hours of unique interaction. The objects span household items, containers, utensils, tools, and electronics. A calibrated array of $32$ RGB cameras records a tabletop workspace at $1920\times1232$ pixels and $30$ frames per second. Duration counts unique interaction time, not the sum over camera views. Recordings include symmetric manipulation, asymmetric coordination, and coordinated multi-object interactions. Participants perform natural object-centric interactions rather than fixed motion templates. These describe the capture inventory; the control formulation tracks two robot hands and one reference object.

We characterize contact as simultaneous or sequential and measure the duration of simultaneous contact. In the analyzed BiHOI recordings, simultaneous bimanual contact accounts for $65.91\%$ of contact-active time (at least one hand in contact). The highest proportion among the analyzed public subsets is $38.57\%$ for H2O, a difference of $27.34$ percentage points. Table~\ref{supp:tab:inventory} reports the selected public training inventory and the measured contact proportions. These subset measurements should not be interpreted as statistics over every recording released by each source.

\begin{table}[t]
\caption{Training inventory and simultaneous bimanual contact. Contact proportions pool frames across analyzed recordings, dividing both-hand-contact frames by frames with at least one hand in contact. The public-only model uses the public subsets listed below; Ours+BiHOI adds all BiHOI sequences.}
\label{supp:tab:inventory}
\centering
\begin{tabular*}{\linewidth}{@{}l@{\extracolsep{\fill}}rr@{}}
\toprule
Source & Training sequences & Simultaneous contact (\%)\\
\midrule
GRAB & 505 & 32.21\\
GigaHands & 592 & 34.74\\
OakInk2 & 449 & 24.97\\
H2O & 127 & 38.57\\
TACO & 172 & 15.69\\
\midrule
Public-data pool & 1,845 & ---\\
BiHOI & 1,575 & 65.91\\
Public-data pool + BiHOI & 3,420 & ---\\
\bottomrule
\end{tabular*}
\end{table}

For the public pool, both hands and the manipulated object must be tracked, and each hand must contact the object during the clip; continuous or simultaneous contact is not an admission requirement. Motions are converted to a common representation consisting of per-hand MANO parameters, an object mesh, and object position and orientation in a gravity-aligned world frame. Retargeting and replay screening then produce robot references. The $53$ unseen-trajectory and $164$ unseen-object evaluation clips are separate from the training counts.

\paragraph{Contact measurement.}
For clip $m$, frame $t$, and hand $h$, let $c_{m,t,i}^h$ denote the reference label at fingertip slot $i\in\{3,6,9,12,15\}$. Define activity $a_{m,t}^h=\ind[\exists i:c_{m,t,i}^h>0.01]$, using the same reference-hand activity rule as training. The reported percentage is
\begin{equation}
 100\,\frac{\sum_{m,t}a_{m,t}^L a_{m,t}^R}
 {\sum_{m,t}\ind[a_{m,t}^L+a_{m,t}^R>0]}.
\end{equation}
Thus the denominator is contact-active time, and aggregation pools frames across clips. It is neither the fraction of all sequence frames nor a mean of per-clip percentages. On the analyzed public pool, this measure is $31.40\%$; using all frames as the denominator instead gives $22.56\%$.

\paragraph{Single-object representation.}
For multi-object recordings, conversion retains one primary object: TACO keeps the tool rather than the target, and OakInk2 selects the main object from its hand-role program annotations. The controller represents the selected object by a fixed mesh and a rigid pose. Articulation and deformation are not modeled explicitly, even when the captured object belongs to those categories.

\subsection{Multi-View Reconstruction and Annotation}
\label{supp:annotation}

\begin{figure}[t]
\centering
\includegraphics[width=\linewidth]{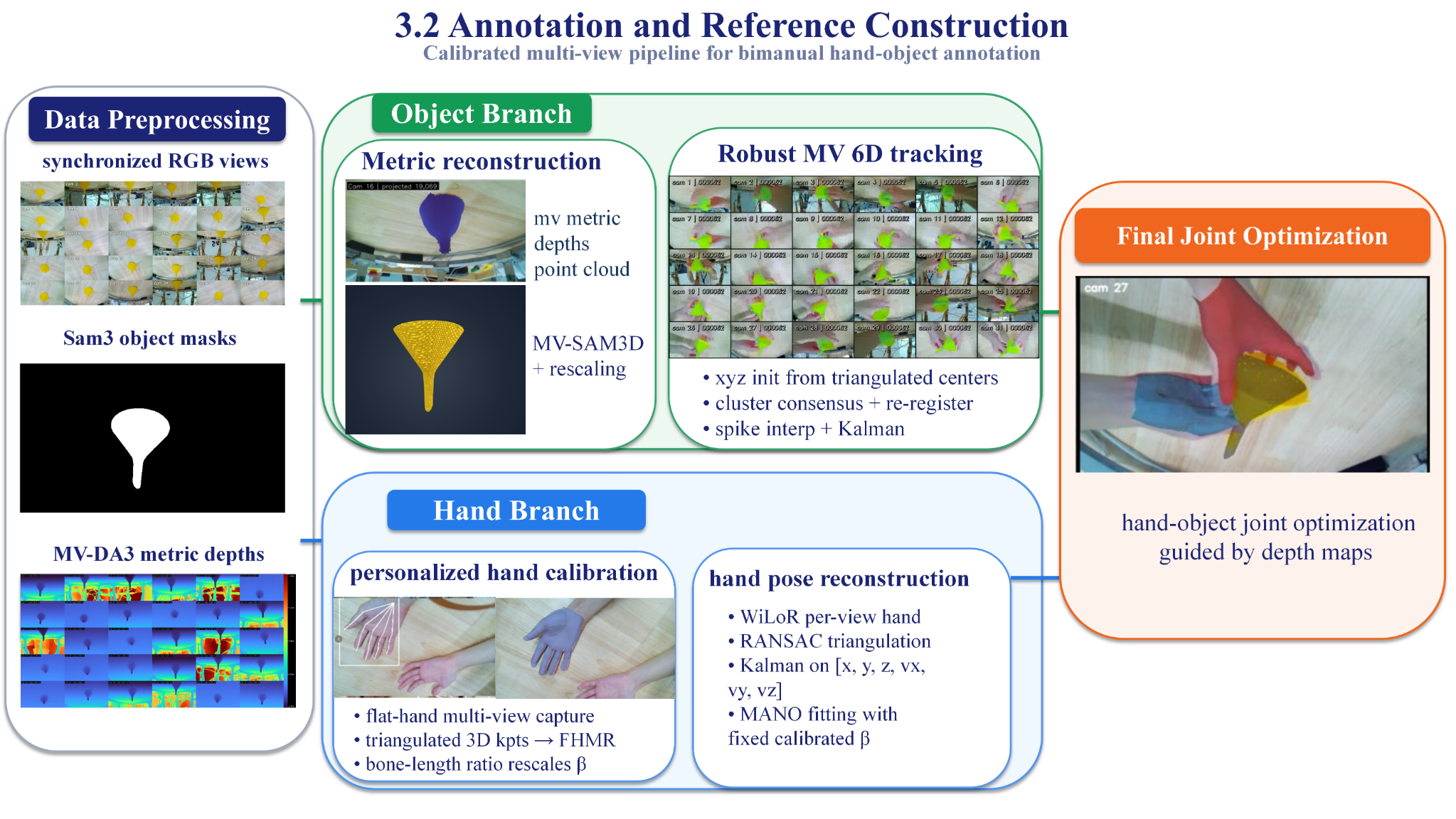}
\caption{\textbf{Reconstruction and annotation pipeline.} Synchronized multi-view RGB observations are processed to obtain SAM3 object masks and metric depth maps. The object branch performs metric reconstruction followed by robust multi-view 6D tracking, while the hand branch performs participant-specific hand calibration and multi-view hand-pose reconstruction. The two branches are finally refined jointly using depth-guided hand--object optimization.}
\label{supp:fig:recon_annotation_pipeline}
\end{figure}

The reconstruction pipeline uses learned image predictions to initialize geometry, then calibrated multi-view observations to recover physical scale and temporal consistency. Its five stages are mask/depth extraction, object reconstruction and tracking, participant-specific hand calibration, hand-motion reconstruction, and joint refinement. Figure~\ref{supp:fig:recon_annotation_pipeline} summarizes the complete calibrated multi-view reconstruction and annotation pipeline.

\paragraph{Masks and metric depth.}
We use SAM3~\citep{carion2025sam3} in two stages. The first frames from all cameras form an ordered cross-camera sequence: a single object prompt initializes consistent masks across views. Each camera video is then tracked temporally using its first-frame mask. At each synchronized time, Depth Anything 3~\citep{lin2025depthanything3} processes the views with calibrated intrinsics and world-to-camera extrinsics. Extrinsic-scale alignment anchors depth to metric camera translations. The resulting depth supports object-scale recovery and subsequent pose refinement.

\paragraph{Metric object reconstruction.}
MV-SAM3D~\citep{li2026mvsam3d} reconstructs a canonical mesh from distributed first-frame views and masks. We back-project masked depth from valid views into the common world frame, reject inconsistent per-camera point-cloud centers, and fuse the remaining observations. To estimate physical scale without aligning an ambiguous canonical orientation, we compare robust principal-component extents of the depth cloud and mesh samples. After removing radial outliers, let $e_k^{\mathrm{depth}}$ and $e_k^{\mathrm{mesh}}$ be their sorted $1$--$99$ percentile extents along the three principal axes. The isotropic scale is
\begin{equation}
 s_{\mathrm{obj}}=\operatorname{median}_{k\in\{1,2,3\}}
 \frac{e_k^{\mathrm{depth}}}{e_k^{\mathrm{mesh}}}.
\end{equation}
Only scale is transferred; the canonical mesh orientation is retained.

\paragraph{Multi-view object tracking.}
We triangulate centers of minimum-area rectangles fitted to object masks to initialize translation, and initialize orientation from the preceding accepted pose. The ten views with the largest visible masks are refined independently with FoundationPose~\citep{wen2024foundationpose}. Candidate poses are transformed into the world frame and grouped by geometric agreement. Two candidates $(R_i,\mathbf t_i)$ and $(R_j,\mathbf t_j)$ are compatible when their rotation geodesic distance $\rotangle(R_i,R_j)<30^\circ$ and $\|\mathbf t_i-\mathbf t_j\|_2<5$\,cm. At least four mutually compatible views are required. Translation is averaged with mask-area weights, while orientation is taken from the largest-mask view in the accepted cluster. Failed consensus triggers re-registration.

The object trajectory undergoes translation spike repair and Kalman filtering, followed by rotation spike repair and filtering on $SO(3)$, the space of three-dimensional rotations. Interpolation is restricted to short spike bursts. We subsequently refine the full trajectory against fused metric depth using alternating iterative closest point correspondence updates. Correspondences farther than $3$\,cm are rejected. For point-to-plane and point-to-point residuals $r_{\mathrm{p2pl}}$ and $r_{\mathrm{p2p}}$, respectively, the geometric contribution is
\begin{equation}
 \mathcal L_{\mathrm{geom}}
 =0.7\,\rho(r_{\mathrm{p2pl}})+0.3\,\rho(r_{\mathrm{p2p}}),
\end{equation}
where $\rho$ is the robust residual penalty. Reconstruction priors and first- and second-order temporal regularization constrain the update. Translation and rotation corrections remain within $3$\,cm and $20^\circ$ of the input trajectory.

\paragraph{Participant-specific hand calibration.}
For each hand, a flat-hand calibration sequence provides metric triangulated positions for a $21$-joint skeleton. An initial MANO shape $\bm\beta_0$~\citep{romero2017mano} is calibrated against these observations. Let $\mathcal B$ contain the $20$ skeleton bones, $\mathbf X_a^{\mathrm{tri}}$ be triangulated joint $a$, and $\mathbf J_a(\bm\beta_0)$ the corresponding model joint. Physical scale is
\begin{equation}
 s_{\mathrm{hand}}=
 \frac{\sum_{(a,b)\in\mathcal B}\|\mathbf X_b^{\mathrm{tri}}-\mathbf X_a^{\mathrm{tri}}\|_2}
 {\sum_{(a,b)\in\mathcal B}\|\mathbf J_b(\bm\beta_0)-\mathbf J_a(\bm\beta_0)\|_2}.
\end{equation}
The explicit scale is retained because MANO shape coefficients are not a global scale parameter. Components requiring a MANO-only shape vector receive an approximate scale-adjusted shape from ridge fitting in centered canonical vertex space. The participant-specific calibration is reused across sequences.

\paragraph{Hand-motion reconstruction.}
WiLoR~\citep{potamias2025wilor} provides $21$ image-space keypoints per hand and view. For each joint, valid two-view pairs generate triangulation hypotheses. Reprojection residuals below $50$ pixels define inliers, with at least three views required. The selected hypothesis is re-triangulated using all inliers. At the reconstruction stage's $10$\,fps sampling rate, each joint is filtered with a constant-velocity state $[x,y,z,v_x,v_y,v_z]^\top$, where $x,y,z$ are coordinates and $v_x,v_y,v_z$ their velocities. Measurement and acceleration-process standard deviations are $0.01$\,m and $0.12$\,m/s$^2$, respectively; missing observations use prediction only. This reconstruction sampling rate is distinct from the $30$\,fps capture rate. MANO fitting combines filtered metric joints, multi-view image keypoints, and calibrated projections, keeping hand shape fixed while estimating pose and global motion.

\paragraph{Joint hand--object refinement.}
The hand and object estimates are refined against shared metric depth. Variables are per-frame hand pose and global transforms, object pose, and one object-scale variable shared by the sequence. The objective is
\begin{equation}
 \mathcal L=10\mathcal L_{\mathrm{SDF}}+
 \mathcal L_{\mathrm{reg}}^h+\mathcal L_{\mathrm{reg}}^o+
 \mathcal L_{\mathrm{smooth}}^h+\mathcal L_{\mathrm{smooth}}^o.
\end{equation}
Here $\mathcal L_{\mathrm{SDF}}$ aligns reconstructed geometry to a signed distance field (SDF) built from depth; $\mathcal L_{\mathrm{reg}}^h$ and $\mathcal L_{\mathrm{reg}}^o$ preserve hand and object initialization, and the two smoothness terms regularize their temporal evolution. The MeshSDF formulation follows HO-Cap~\citep{wang2024hocap}, using $32$ depth views in the interaction region and $3$\,mm voxels. Optimization first uses priors and smoothness, then activates SDF alignment, and finally refines object scale with poses fixed. It does not include explicit contact, penetration, collision, or snapping losses.

\paragraph{Export representation.}
The standardized export uses a right-handed, $+Z$-up world frame with its recorded $4\times4$ transform from the reconstruction frame. The tabletop defines $z=0$, estimated from the fifth percentile of the object bottom. MANO uses 45 axis--angle pose coordinates without PCA, with a flat-hand mean and no added mean-pose offset. Object vertices retain their canonical coordinates without recentering, axis flips, or mirroring.

\subsection{Kinematic Retargeting and Replay Screening}
\label{supp:retarget}

Retargeting follows position-based kinematic fitting~\citep{handa2020dexpilot,qin2023anyteleop,liu2025dextrack}. For each frame and hand, fifteen target entries are fitted to corresponding Wuji-hand links: five MANO fingertips (indices $4,8,12,16,20$), five additional finger keypoints ($3,7,11,15,19$), and the wrist (index $0$) repeated five times. The fit uses SmoothL1 position loss with transition scale $0.02$\,m and NLopt optimization with robot joint limits as hard bounds. The implementation uses a joint-limit tolerance of $10^{-3}$, initializes from the preceding frame's solution, and uses inter-frame regularization parameter \texttt{norm\_delta}$=4\times10^{-3}$.

At conversion, source motion is interpreted at $10$\,fps and converted to $30$\,fps control references (\texttt{SOURCE\_FPS=10}, \texttt{FPS\_OUT=30}, \texttt{RESAMPLE\_STRIDE=3}). The source visualization videos and per-clip provenance determine this timing; the export metadata's nominal $30$\,fps field is not used as the source rate. Pre-retargeting cleanup uses a $0.10$\,m per-frame displacement threshold at $30$\,fps, at most three despiking passes, and nearest-neighbor filling. These settings belong to conversion cleanup, not reconstruction filtering.

Trajectory cleanup occurs before retargeting. The retargeter's low-pass coefficient is $1$, so it performs no additional low-pass smoothing. Wrist translation is computed as the MANO global translation plus its shape-dependent root joint, $\mathbf t+\mathbf J_0(\bm\beta)$; a fixed frame correction aligns global orientation with the robot convention. The output pairs robot wrist/finger configurations with the reconstructed object trajectory.

Open-loop replay with zero policy correction classifies each candidate reference into six outcomes: immediate object loss (A), object ejection (B), contact without lifting (C), partial execution (D), tracking with an incorrect release or ending (E), and complete execution (F). Categories A and B are excluded, whereas C--F remain available for closed-loop training. Category C is retained because failure to lift under reference-only replay does not rule out recovery by feedback control. Screening determines reference eligibility; it is distinct from the verification used to admit executable demonstrations.

\section{Method and Training Details}
\label{supp:method}
\setcounter{equation}{0}

\subsection{Controller Interface}
\label{supp:controller}

A reference clip $m$ contains robot configurations $\hat{\mathbf q}_t^h\in\R^{26}$ for hand $h\in\{L,R\}$, object position $\hat{\mathbf p}_t^o\in\R^3$, and object rotation $\hat R_t^o\in SO(3)$ at control step $t$. Hats denote references; quantities without hats denote simulated states. Each hand has six wrist coordinates and twenty finger joints. The shared policy $\pi_\theta(\mathbf a_t\mid\mathbf o_t)$, with parameters $\theta$, receives observation $\mathbf o_t$ and produces a $52$-dimensional action $\mathbf a_t$ for both hands.

The $3{,}016$-dimensional observation combines $2{,}884$ base features with a $132$-dimensional object descriptor: $128$ basis-point-set features~\citep{prokudin2019bps}, scale, and centroid. Per-hand blocks contain proprioception, object pose and velocity, upcoming reference targets, state errors, and fingertip-to-object distance vectors in each wrist heading frame. Hand identity and left-hand reward weight are included. Forty contact-force channels are set to zero; contact forces contribute to reward supervision, not policy input.

Concatenating both reference configurations as $\hat{\mathbf q}_{t+1}$, the proportional--derivative (PD) target is
\begin{equation}
 \mathbf q_t^{\mathrm{pd}}=\operatorname{clamp}
 (\hat{\mathbf q}_{t+1}+\mathbf s\odot\mathbf a_t,
 \mathbf q_{\min},\mathbf q_{\max}),\qquad \mathbf a_t\in[-1,1]^{52}.
\end{equation}
The vector $\mathbf s$ sets coordinate-specific correction bounds, $\odot$ denotes element-wise multiplication, and $\mathbf q_{\min},\mathbf q_{\max}$ are coordinate limits. Bounds are $0.1$\,m for wrist translation, $1.0$\,rad for wrist rotation, and $0.15$\,rad for finger joints. Corrections are applied around the next reference at every step and do not accumulate. This interface is inherited from reference-tracking approaches~\citep{liu2025dextrack,xu2025dexplore}.

\subsection{Reward Definition and Coefficients}
\label{supp:reward}

We combine hand tracking, object motion, per-hand contact, and contact maintenance:
\begin{equation}
 r_t=(1+w_L)\big[\beta_{\mathrm{hand}}r_t^{\mathrm{hand}}
 +\beta_{\mathrm{obj}}r_t^{\mathrm{obj}}
 +\beta_{\mathrm{cont}}r_t^{\mathrm{cont}}
 +\beta_{\mathrm{hold}}r_t^{\mathrm{hold}}\big].
\end{equation}
The fixed hand weights are $w_L=0.5$ and $w_R=1$; adaptive hand weighting is disabled. The $\beta$ coefficients sum to one. The inherited hand-tracking term uses Dexplore's per-hand multiplicative chain for wrist, finger-joint, and keypoint agreement~\citep{xu2025dexplore}, with the released subterm coefficients unchanged. Its two per-hand values are combined as $(r_t^{\mathrm{hand},R}+w_Lr_t^{\mathrm{hand},L})/(1+w_L)$. Object-position and object-rotation factors within this hand chain have zero weights (\texttt{reward.op=reward.or=0}), avoiding double counting with the separate object reward.

For object linear velocity $\mathbf v_t^o$ and reference velocity $\hat{\mathbf v}_t^o$ from finite differences, the object reward is
\begin{equation}
 r_t^{\mathrm{obj}}=
 e^{-\kappa_p\|\mathbf p_t^o-\hat{\mathbf p}_t^o\|_2}
 e^{-\kappa_R\rotangle(R_t^o,\hat R_t^o)}
 e^{-\kappa_v\|\mathbf v_t^o-\hat{\mathbf v}_t^o\|_2}.
\end{equation}
The implemented object reward multiplies the three factors and has no $\eta$ mixture weights. Joint position, orientation, and velocity agreement is rewarded; the velocity factor penalizes holding a moving reference stationary. This term is independent of contact gating. The positive coefficients $(\kappa_p,\kappa_R,\kappa_v)=(9,2,14)$ control the three error sensitivities. The full reward still adds the hand, object, contact, and hold terms.

The reference-required set $\mathcal C_t^h$ uses the five fingertip slots $[3,6,9,12,15]$ within reference contact columns $[206{:}222]$ for the left hand and $[222{:}238]$ for the right. A fingertip is required when its channel exceeds $0.01$; hand activity uses an element-wise logical OR over these tests, not a sum of signed labels. This reference mask is distinct from simulated contact. For fingertip $i$, surface distance $\delta_{i,t}^h$ and simulated net force magnitude $f_{i,t}^h$ define a contact proxy
\begin{equation}
 \alpha_{i,t}^h=\ind[f_{i,t}^h>f_{\min}]\ind[\delta_{i,t}^h<\delta_{\max}].
\end{equation}
For nonempty $\mathcal C_t^h$, proximity and instantaneous required-contact coverage give
\begin{equation}
\begin{aligned}
 r_t^{\mathrm{cont},h}&=\frac{1}{|\mathcal C_t^h|}\sum_{i\in\mathcal C_t^h}
 \left[\zeta e^{-\kappa_c\delta_{i,t}^h}+(1-\zeta)\alpha_{i,t}^h\right],\\
 r_t^{\mathrm{hold},h}&=\frac{1}{|\mathcal C_t^h|}\sum_{i\in\mathcal C_t^h}\alpha_{i,t}^h.
\end{aligned}
\end{equation}
The set of engaged hands is $\mathcal E_t=\{h:\mathcal C_t^h\ne\varnothing\}$. Each contact-related term $u\in\{\mathrm{cont},\mathrm{hold}\}$ is aggregated as
\begin{equation}
 r_t^u=\begin{cases}
 \displaystyle\frac{\sum_{h\in\mathcal E_t}w_h r_t^{u,h}}{\sum_{h\in\mathcal E_t}w_h},&\mathcal E_t\ne\varnothing,\\[3pt]
 0,&\mathcal E_t=\varnothing.
 \end{cases}
\end{equation}
Thus an inactive hand contributes no contact reward, and an unsuccessful engaged hand does not multiply away the other hand's signal. The force--distance rule remains a proxy because net force does not identify the collision partner. Contact maintenance measures instantaneous coverage, not a no-slip condition, allowing valid sliding and rolling. Proximity guides fingertips before contact; the hold term evaluates required-contact coverage throughout the reference contact interval.

\begin{table}[t]
\caption{Reward and generalist-training settings. Distances in reward exponents use metres, rotations radians, and velocities metres per second.}
\label{supp:tab:settings}
\centering\small
\begin{tabular*}{\linewidth}{@{}l@{\extracolsep{\fill}}l@{}}
\toprule
Parameter & Setting\\
\midrule
$(\beta_{\mathrm{hand}},\beta_{\mathrm{obj}},\beta_{\mathrm{cont}},\beta_{\mathrm{hold}})$ & $(0.25,0.45,0.15,0.15)$\\
$(\kappa_p,\kappa_R,\kappa_v)$ & $(9,2,14)$\\
$(f_{\min},\delta_{\max})$ & $(0.5\,\mathrm N,0.03\,\mathrm m)$\\
$(\kappa_c,\zeta)$ & $(30\,\mathrm m^{-1},0.5)$\\
Parallel simulation environments & 8,192\\
BC coefficient / pairs per minibatch & 50 / 4,096\\
Recovery / perturbation curriculum & Disabled\\
Fixed hand weights $(w_L,w_R)$ & $(0.5,1)$\\
\bottomrule
\end{tabular*}
\end{table}

\subsection{Clip-Centered Advantages and PPO+BC}
\label{supp:cca}

We use proximal policy optimization (PPO)~\citep{schulman2017ppo} with generalized advantage estimation (GAE)~\citep{schulman2016gae}, and add behavioral cloning (BC) from verified demonstrations. Let $V_\phi(\mathbf o_t)$ denote the learned value estimate, $\gamma$ the discount factor, and $\lambda_{\mathrm{GAE}}$ the temporal trace parameter. Writing $b_t$ for the bootstrap-validity mask, the temporal-difference residual is
\begin{equation}
 \Delta_t=r_t+\gamma b_t V_\phi(\mathbf o_{t+1})-V_\phi(\mathbf o_t).
\end{equation}
GAE accumulates these residuals with weights $(\gamma\lambda_{\mathrm{GAE}})^\ell$, stopping the trace at episode boundaries. Bootstrap and timeout handling follow the base implementation. This temporal estimate is computed before any clip-level adjustment.

For an on-policy batch $\mathcal B$, transition index $j$, and reference identity $m(j)$, define $\mathcal G_m=\{j\in\mathcal B:m(j)=m\}$. CCA uses
\begin{equation}
 \bar A_m=\frac{1}{|\mathcal G_m|}\sum_{j\in\mathcal G_m}A_j^{\mathrm{GAE}},\qquad
 \widetilde A_j=A_j^{\mathrm{GAE}}-\bar A_{m(j)}.
\end{equation}
Each nonempty group has zero empirical mean after centering. For two transitions $j,k$ from the same clip, $\widetilde A_j-\widetilde A_k=A_j^{\mathrm{GAE}}-A_k^{\mathrm{GAE}}$. These identities explain the intended behavior: remove a common reference-dependent offset while preserving within-reference differences. CCA is a mean shift, not division by a clip standard deviation, and it retains the learned critic and temporal GAE calculation. Its empirical centering changes the actor update; the algebraic identities alone do not establish an unbiased policy-gradient estimator.

Let $\theta_{\mathrm{old}}$ denote the policy that collected $\mathcal B$, $\epsilon$ the PPO clipping radius, and
$\rho_j(\theta)=\pi_\theta(\mathbf a_j\mid\mathbf o_j)/\pi_{\theta_{\mathrm{old}}}(\mathbf a_j\mid\mathbf o_j)$.
The actor loss is
\begin{equation}
\begin{split}
 \mathcal L_{\mathrm{PPO}}=-\E_{j\sim\mathcal B}\big[\min\big(&\rho_j(\theta)\widetilde A_j,\\[-2pt]
 &\operatorname{clip}(\rho_j(\theta),1-\epsilon,1+\epsilon)\widetilde A_j\big)\big].
\end{split}
\end{equation}
For demonstration library $\mathcal D$ and policy mean $\mu_\theta$, the combined objective is
\begin{equation}
 \mathcal L_{\mathrm{actor}}=\mathcal L_{\mathrm{PPO}}+
 \lambda_{\mathrm{BC}}\E_{(\mathbf o^D,\mathbf a^D)\sim\mathcal D}
 \|\mu_\theta(\mathbf o^D)-\mathbf a^D\|_2^2,
 \qquad \lambda_{\mathrm{BC}}=50.
\end{equation}
The superscript $D$ denotes a stored observation--action pair. Demonstrations use a separate sampling budget and do not enter the PPO probability ratio or GAE. The critic retains its value-regression objective.

\subsection{Demonstration Collection, Sampling, and Compatibility}
\label{supp:demos}

Following DexTrack's reinforcement--imitation loop~\citep{liu2025dextrack}, specialist training covers all references retained by replay screening. Each trajectory-specific controller seeks an executable realization of its reference; successful rollouts are exported and replay-verified before admission to the demonstration library. The shared controller is periodically evaluated on the training pool, and its verified successes supplement the library for subsequent PPO+BC updates. Held-out references are excluded throughout.

Verification checks object displacement and, when required by the reference, lifting. An exported rollout must succeed again on replay. This admission procedure selects executable supervision and is distinct from the trajectory-fidelity metrics in Section~\ref{supp:metrics}. Each BC minibatch samples 4,096 observation--action pairs independently of the PPO minibatch.

A versioned action/reference contract records the action parameterization, reference version, and retargeting fingerprint. Changes to this contract invalidate incompatible stored actions.

\subsection{Optimization and Simulation Settings}
\label{supp:training}

The generalist recipe uses Isaac Gym~\citep{makoviychuk2021isaacgym} with 8,192 environments. Actor and critic use separate multilayer perceptrons with widths $[1024,1024,1024,512]$ and ReLU activations. The Gaussian policy uses a fixed, non-learned scale parameter configured by \texttt{sigma\_init=-2.9}. Input, value, and advantage normalization and value bootstrapping are enabled.

\begin{table}[t]
\caption{Optimization and simulation configuration.}
\label{supp:tab:optimization}
\centering\small
\begin{tabular*}{\linewidth}{@{}l@{\extracolsep{\fill}}l@{}}
\toprule
Setting & Value\\
\midrule
Optimizer / learning rate / schedule & Adam / $2\times10^{-5}$ / constant\\
Rollout horizon / PPO minibatch & 64 / 16,384\\
PPO passes per rollout / clipping radius & 6 / 0.2\\
Discount / GAE trace parameter & 0.99 / 0.95\\
Value / entropy / action-bound loss coefficients & 5 / 0 / 10\\
Value clipping / gradient truncation & Disabled / disabled\\
Configured gradient norm / seed & 1.0 / 42\\
Maximum episode steps & 2,000\\
Control-frequency inverse / simulation substeps & 2 / 4\\
Object density / hand friction & 400 / 2.0\\
Linear / angular object-velocity caps & 20 / 50\\
Reset angular-velocity cap & 20\\
Pre-grasp initialization & 0.9\\
Reference feed-forward / deployment mask & Enabled / enabled\\
Reference-correction bounds (wrist trans./rot.; fingers) & $0.1$\,m / $1.0$\,rad / $0.15$\,rad\\
BC coefficient / demonstration batch & 50 / 4,096\\
Recovery / object-perturbation curriculum & Disabled\\
Environment transitions per rollout iteration & $8{,}192\times64=524{,}288$\\
\bottomrule
\end{tabular*}
\end{table}

Demonstration supervision uses $\lambda_{\mathrm{BC}}=50$ with a separate minibatch budget. Adaptive recovery sampling and object-position perturbation curricula are disabled. Checkpoints are saved every 25 training iterations.

\section{Experimental Protocols and Additional Details}
\label{supp:exp}
\setcounter{equation}{0}

\subsection{Evaluation Splits}
\label{supp:splits}

The public-only model uses $1{,}845$ references and Ours+BiHOI uses $3{,}420$, including all $1{,}575$ BiHOI sequences. Evaluation comprises $53$ new trajectories of training objects (\textsc{unseen-traj}) and $164$ clips involving held-out objects (\textsc{unseen-obj}). Neither split contributes to policy optimization, specialist training, or demonstration collection. Each method uses one checkpoint across both splits without fine-tuning.

The main DONIC results use a single frozen checkpoint. Evaluation uses the policy mean and starts clips from their first frame. Neither evaluation split supplies training transitions or demonstrations.

\subsection{Bimanual Tracking Metrics}
\label{supp:metrics}

We extend AdaDexTrack's trajectory-fidelity evaluation~\citep{adalibieke2026adadextrack} to require both hands to satisfy the tracking criteria. Let $\mathcal M$ be the evaluated clips, $T_m$ the number of evaluated frames in clip $m$, and $h\in\{L,R\}$ a hand. The following definitions apply to each frame; clip indices are omitted when unambiguous.

\paragraph{Object, wrist, and finger errors.}
Object translation error $e_{p,t}^o$ is the Euclidean distance between simulated and reference positions. For unit object-orientation quaternions $\mathbf u_t^o$ and $\hat{\mathbf u}_t^o$, rotation error is
\begin{equation}
 e_{R,t}^o=2\arccos\left(|\langle\mathbf u_t^o,\hat{\mathbf u}_t^o\rangle|\right).
\end{equation}
The absolute inner product accounts for the quaternion double cover, giving an angle in $[0,\pi]$.

For each hand, the same robot Unified Robot Description Format (URDF) forward kinematics maps simulated and retargeted reference configurations to wrist poses $(\mathbf p_t^h,R_t^h)$ and $(\hat{\mathbf p}_t^h,\hat R_t^h)$. The wrist errors are
\begin{equation}
 e_{p,t}^h=\|\mathbf p_t^h-\hat{\mathbf p}_t^h\|_2,\qquad
 e_{R,t}^h=\arccos\!\left(\frac{\operatorname{tr}((R_t^h)^\top\hat R_t^h)-1}{2}\right).
\end{equation}
The source human wrist is not used in this comparison: the metric measures tracking after retargeting, rather than adding human-to-robot retargeting mismatch. With finger-joint angles $q_{t,k}^h$ and $\hat q_{t,k}^h$, finger error is
\begin{equation}
 e_{f,t}^h=\frac{1}{20}\sum_{k=1}^{20}|q_{t,k}^h-\hat q_{t,k}^h|.
\end{equation}
The six virtual wrist coordinates are excluded. All angular errors and thresholds are computed in radians; translation errors are expressed in centimetres for reporting and thresholding.

\paragraph{Aggregation of reported errors.}
For any frame error $e$, write $\bar e_m=T_m^{-1}\sum_t e_{m,t}$ for its trajectory mean. The reported object metrics are
\begin{equation}
 T_o=\frac{1}{|\mathcal M|}\sum_m\bar e_{p,m}^o,\qquad
 R_o=\frac{1}{|\mathcal M|}\sum_m\bar e_{R,m}^o.
\end{equation}
Hand metrics first take the worse of the two per-hand trajectory means, then average over clips:
\begin{equation}
 (T_h,R_h,E_{\mathrm{finger}})=\frac{1}{|\mathcal M|}\sum_m
 \left(\max_h\bar e_{p,m}^h,\max_h\bar e_{R,m}^h,\max_h\bar e_{f,m}^h\right).
\end{equation}
The worse hand can differ by metric. In particular, this is not an average across hands or a framewise maximum before temporal averaging.

\paragraph{Success rate and completion.}
For object errors $x_p^o,x_R^o$ and hand errors $x_p^h,x_R^h,x_f^h$, define
\begin{equation}
 \mathcal K(x^o,x^h)=
 \ind\!\left[\frac{x_p^o+x_p^h}{2}\le10\,{\rm cm}\right]
 \ind\!\left[\frac{x_R^o+x_R^h}{2}\le\frac{\pi}{3}\right]
 \ind\!\left[x_f^h\le\frac{2\pi}{9}\right].
\end{equation}
The angular thresholds $\pi/3$ and $2\pi/9$ correspond to $60^\circ$ and $40^\circ$. Let $\bar e_m^o$ and $\bar e_m^h$ collect the relevant trajectory means. Mean-error success rate is
\begin{equation}
 \mathrm{SR}=\frac{100}{|\mathcal M|}\sum_m
 \prod_{h\in\{L,R\}}\mathcal K(\bar e_m^o,\bar e_m^h).
\end{equation}
The product implements a logical AND: both hands must independently pass using their own trajectory means. Completion instead applies the same tests separately at every frame:
\begin{equation}
 \mathrm{Comp}=\frac{100}{|\mathcal M|}\sum_m\frac{1}{T_m}
 \sum_t\prod_{h\in\{L,R\}}\mathcal K(e_{m,t}^o,e_{m,t}^h).
\end{equation}
A later valid frame contributes even after an earlier violation. Comp therefore measures per-frame satisfaction, not the length of the prefix before the first failure. Both metrics weight trajectories equally rather than weighting longer clips more heavily.

\subsection{Baseline Implementations and Comparison Scope}
\label{supp:baselines}

\paragraph{DexTrack-bi.}
We use DexTrack's released specialist--distillation pipeline~\citep{liu2025dextrack}, adapting its observation and action interfaces to two Wuji hands. Trajectory-specific controllers produce demonstrations for generalist distillation, and evaluation follows the released per-trajectory protocol.

\paragraph{Dexplore-bi.}
We run the released Dexplore code~\citep{xu2025dexplore} with two-hand observation and action interfaces, retaining its original reward and reference-scoped exploration.

\paragraph{Evaluation settings.}
All methods use Isaac Gym, two Wuji hands, the same public training clips, and the same held-out split identities. Tracking errors are computed using the bimanual metrics defined in Section~\ref{supp:metrics}. All 164 unseen-object and 53 unseen-trajectory clips are included.

\subsection{Ablation Definitions and Additional Data}
\label{supp:ablations}

The learning-signal ablations in the main table examine CCA and object-centered supervision. The w/o CCA variant removes the clip-wise mean subtraction after GAE. The w/o contact gate variant disables \texttt{DEXPLORE\_CG\_REWARD}, replacing the simulated force-and-distance contact proxy by the force-only test $\ind[f_{i,t}^h>0.5]$. It removes the $3$\,cm contact-ownership condition, not the object-motion reward. Reward coefficients $\beta$ and fixed hand weights are unchanged; they are not renormalized. Ours+BiHOI tests the effect of adding all $1{,}575$ captured sequences to the public pool while retaining the same public evaluation splits.

\paragraph{Ablation outcomes.}
Removing either learning signal lowers both SR and Comp on both splits. CCA removal also increases all five tracking errors; removing the contact gate improves some individual errors but reduces joint bimanual success. Adding BiHOI improves all seven metrics on both splits, reaching SR of $88.68\%$ and $59.76\%$ on unseen trajectories and objects, respectively. Unseen-object translation error falls from $15.74$ to $13.96$\,cm.

The learning-signal ablations retain the full model's training data and optimization settings, changing only the indicated component. Ours+BiHOI uses the same training recipe with the expanded reference pool.

\subsection{Qualitative Comparisons}
\label{supp:qualitative}

\begin{figure}[p]
\centering
\includegraphics[width=\linewidth,height=0.90\textheight,keepaspectratio]{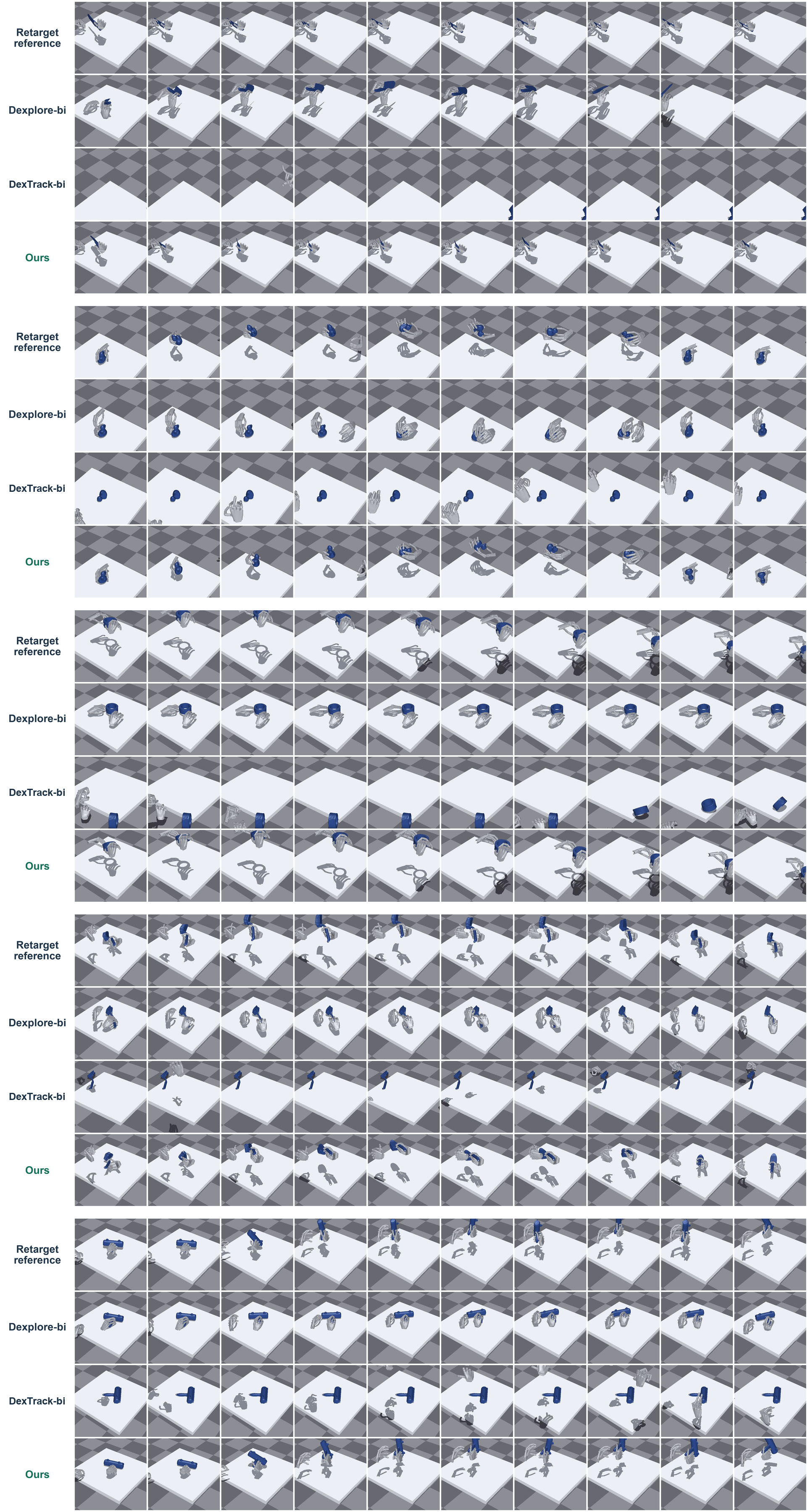}
\caption{\textbf{Qualitative bimanual tracking comparisons.} Rows in each group show the reference, Dexplore-bi, DexTrack-bi, and \method\ (Ours); time runs left to right.}
\label{supp:fig:all_comparisons}
\end{figure}

Figure~\ref{supp:fig:all_comparisons} compares five bimanual manipulation sequences. Across these examples, \method\ more closely reproduces reference hand--object coordination and object motion, whereas the baselines exhibit limited object motion or loss of hand--object alignment.

\subsection{Limitations}
\label{supp:limitations}

The quantitative experiments evaluate reference-conditioned, single-rigid-object tracking in simulation. Unconstrained goal-conditioned manipulation and held-out operation-category generalization are outside this evaluation. The capture inventory includes broader object types than the rigid-object representation described in Section~\ref{supp:inventory}. Contact supervision uses a net-force/proximity proxy, and reconstruction has no explicit contact or collision loss.

\end{document}